\documentclass[lettersize,journal]{IEEEtran}
\usepackage{amsmath,amsfonts}
\usepackage{algorithmic}
\usepackage{algorithm}
\usepackage{array}
\usepackage[caption=false,font=normalsize,labelfont=sf,textfont=sf]{subfig}
\usepackage{textcomp}
\usepackage{stfloats}
\usepackage{url}
\usepackage{verbatim}
\usepackage{graphicx}
\usepackage{cite}

\usepackage{multirow}
\usepackage{arydshln}  
\usepackage{pifont}  

\usepackage{booktabs}

\usepackage{hyperref}

\begin{document}

\title{MLCR: Multi-Level Cue Refinement for Long-Term Multimodal Action Quality Assessment}

\author{Qiqi Li, Pengfei Wang, Hongyu Chen, and Nenggan Zheng,~\IEEEmembership{Senior Member,~IEEE}
\thanks{Qiqi Li and Hongyu Chen are with Qiushi Academy for Advanced Studies (QAAS), Zhejiang University, Hangzhou, Zhejiang, 310007, China, and also with College of Computer Science and Technology, Zhejiang University, Hangzhou, Zhejiang, 310007, China. (e-mail: qiqili1023@zju.edu.cn, c\_hy@zju.edu.cn).}
\thanks{Pengfei Wang is with School of Software Technology, Zhejiang University, Hangzhou, Zhejiang, 310007, China. (e-mail: pfei@zju.edu.cn).}
\thanks{Nenggan Zheng is with Qiushi Academy for Advanced Studies (QAAS), Zhejiang University, Hangzhou, Zhejiang, 310007, China, with the State Key Lab of Brain-Machine Intelligence, Hangzhou, Zhejiang, 310007, China, with Collaborative Innovation Center for Artificial Intelligence by MOE and Zhejiang Provincial Government (ZJU), Hangzhou, Zhejiang, 310007, China, and also with the Zhejiang Lab, Hangzhou, Zhejiang, 311121, China. (e-mail: zng@cs.zju.edu.cn).
\textit{(Corresponding author: Nenggan Zheng.)}
}
}

\markboth{Journal of \LaTeX\ Class Files,~Vol.~14, No.~8, August~2021}%
{Shell \MakeLowercase{\textit{et al.}}: A Sample Article Using IEEEtran.cls for IEEE Journals}


\maketitle

\begin{abstract}

Long-term multimodal action quality assessment (AQA) evaluates action execution within several-minute audiovisual sequences by mining discriminative quality cues for score prediction. However, existing multimodal methods uniformly model entire sequences via a single temporal encoder followed by direct feature alignment or concatenation. This causes critical cues easily obscured by global trends, weakened by modal redundancy, and distorted during one-shot score mapping. We reformulate long-term multimodal AQA as a quality cue organization problem, where score-relevant cues should be progressively identified, retrieved, and accumulated from heterogeneous modalities and long temporal sequences. Based on this perspective, we develop MLCR, a multi-level quality cue refinement framework that organizes quality evidence at three levels including intra-modal representation, cross-modal interaction, and stage-wise aggregation. At the intra-modal level, the intra-modal decoupling encoder (IMDE) preserves identity semantics while refining global temporal context and local frequency details, thereby providing structured candidate cues of different granularities. At the cross-modal level, the cross-modal dynamic complementarity-aware retrieval (CMDCR) module retrieves incremental evidence conditioned on the evolving fused state, reducing redundant modality responses while emphasizing modality-specific quality cues. At the aggregation level, the stage-wise multimodal integration (SMI) block decomposes one-shot score mapping into progressive cue accumulation and residual correction, incrementally consolidating intra-modal and cross-modal cues for quality representation refinement. Extensive experiments on the Rhythmic Gymnastics and Fis-V datasets show that MLCR achieves the best or second-best performance in both Spearman’s correlation and prediction error, demonstrating its effectiveness and robustness.

\end{abstract}


\section{Introduction}

\IEEEPARstart{A}{ction} quality assessment (AQA) aims to infer expert-level judgments from spatiotemporal audiovisual signals, with important applications in competitive sports scoring\cite{wang2021tsa,yin2026decade}. Existing methods have evolved from handcrafted features to deep visual backbones such as C3D\cite{tran2015learning}, I3D\cite{carreira2017quo}, and VST\cite{liu2022video}, and further to graph-based\cite{liu2025adaptive,zhou2023hierarchical,pan2019action} models. However, most of these methods build upon a simplified global sequence regression paradigm, compressing an entire performance into a single global representation for score prediction\cite{dong2026uil}. This paradigm works well for several-second, single-modality benchmarks such as diving~\cite{ding20242m,tang2020uncertainty}, but struggles with several-minute long-term performances such as rhythmic gymnastics and figure skating\cite{zhou2025phi}, where quality-relevant cues are distributed across complex temporal and cross-modal interactions. Scoring-relevant cues are sparsely distributed across performance phases and jointly influenced by technical execution and rhythmic synchronization, surpassing the capacity of conventional global regression paradigms.

The key challenge of long-term multimodal AQA is therefore not merely to extract stronger features, but to organize quality-relevant cues from heterogeneous modalities and long temporal sequences. Unlike short-term actions, music in long-term actions serves not merely as background but as a substantive scoring component. Audio carries rich temporal structural cues, including action–music synchronization, beat dynamics, and key action boundaries~\cite{wang2025beats}, providing complementary evidence beyond visual information for quality assessment. Accordingly, audio-visual AQA methods~\cite{zeng2024multimodal,xia2023skating} have emerged as a promising direction for long-term scenarios. However, visual and audio modalities originate from different semantic spaces, and their quality-relevant cues are temporally non-uniform and exhibit both complementarity and redundancy. Thus, a long-term multimodal AQA model should progressively identify, retrieve, and accumulate discriminative quality cues throughout the entire performance.

This requirement is also consistent with the logic of expert evaluation. In disciplines such as rhythmic gymnastics and figure skating, judgments involve scoring criteria and iterative review\cite{majeedi2024rica,ji2023localization,matsuyama2023iris}. The technical criterion assesses the difficulty and execution of key actions including jumps, rotations, lifts, throws, landing stability, and apparatus control~\cite{rai2025rubric}. The artistic criterion evaluates choreography diversity, music integration, rhythm synchronization, and overall fluency~\cite{wang2026beatscore}. Meanwhile, expert evaluation often revisits critical actions and local anomalies before assigning a final score, rather than relying on a single holistic impression. These evaluation mechanisms suggest that long-term AQA is fundamentally a process of progressive organization and accumulation of quality cues. Therefore, models should capture both global trends and discriminative local details, establish effective audio-visual complementarity, and support progressive reasoning and refinement. These requirements motivate quality cue organization at three levels: intra-modal representation, cross-modal interaction, and stage-wise aggregation.

Although various audio-visual AQA methods have been proposed, their ability to organize quality cues remains limited. Early work\cite{xia2023skating} validated the effectiveness of joint audio-visual modeling by employing memory recurrent units to mix clip-level features and accumulate long-range temporal information, but direct fusion of heterogeneous modality features introduces inter-modal interference that weakens the discriminative cues of individual modalities. Subsequent work\cite{zeng2024multimodal} incorporated a mixed-modality branch alongside unimodal branches to mitigate this interference, but overlooked intra-modal multi-scale quality cues entangled within a shared representation space. In addition, other works\cite{wang2025attention} enhanced temporal audio-visual interaction through attention-based cross-modal alignment, but overlooked the computational cost caused by redundant information. To improve computational efficiency, some studies\cite{wang2025learning} adopted selective state space models (SSMs)\cite{S4gu2022ICLR,gu2024mamba} for long-term sequence modeling. However, the recurrent compression mechanism of SSM systematically attenuates high-frequency local cues, weakening the representation of sparse quality details at critical moments. Furthermore, several studies\cite{li2026techcoach,xu2025language,du2024learning,zhang2024narrative} leverage domain knowledge to guide audio-visual learning, but still treat long-term performances as relatively homogeneous processes, lacking explicit modeling of the dynamic evolution of quality cues across scoring stages.

Despite these efforts, existing audio-visual AQA methods still face challenges in organizing quality cues at the intra-modal, cross-modal, and stage-wise aggregation levels. First, quality cues within each modality are entangled. Global cues of fluency and rhythm consistency require long-range context, whereas local cues of jitter, beat deviations, and execution errors are sparse perturbations but strongly affect the final score. Forcing both cue types to share a unified temporal encoder not only risks burying critical local cues in global averaging, but also incurs unnecessary computational overhead. Second, cross-modal quality cues are both complementary and redundant. RGB, optical flow, and audio provide evidence from distinct semantic spaces, where useful information tends to be complementary rather than merely similar\cite{lin2025enhancing}. Nevertheless, prevailing similarity-driven cross-modal attention systematically suppresses complementary signals, while naive concatenation or weighted summation introduces redundant interference, induces bias toward dominant modalities, and amplifies noise from weaker ones. Third, one-shot feature fusion is insufficient for complex long-term quality mapping. Expert scoring is inherently a progressive reasoning process from estimation to refinement, whereas existing methods compress it into single-stage modeling and lack cue-updating mechanism, causing biases in early representations to remain uncorrected and propagate directly to the final prediction.

To address these issues, we propose MLCR, a multi-level quality cue refinement framework for long-term multimodal AQA. Rather than simplifying AQA as holistic sequence regression, MLCR organizes score-relevant evidence through intra-modal cue structuring, cross-modal complementary retrieval, and stage-wise cue aggregation. At the \textbf{intra-modal representation} level, the intra-modal decoupling encoder (IMDE) separates each modality into identity-preserved and quality-enhanced components, preserving stable modality semantics while refining global temporal context and local frequency details within a unified representation. This alleviates cue entanglement and provides structured modality-specific candidate cues at different granularities. At the \textbf{cross-modal interaction} level, we propose a cross-modal dynamic complementarity-aware retrieval (CMDCR) module to address the coexistence of complementarity and redundancy among multimodal cues. CMDCR conditions modality retrieval on the evolving fused representation, emphasizing incremental modality-specific quality cues not sufficiently covered by previous fusion stages while suppressing redundant responses. At the \textbf{stage-wise aggregation} level, we adopt a stage-wise cue fusion strategy instead of one-shot mapping. Through the stage-wise multimodal integration (SMI) block, modality-specific intra-modal cues and cross-modal complementary cues are progressively coordinated, allowing quality evidence from different modalities and temporal scales to be accumulated into a refined fused representation. Through this multi-level cue organization mechanism, MLCR constructs long-term multimodal quality representations that are better aligned with the logic of expert scoring. The contributions are summarized as follows:

\begin{itemize}

	\item This work introduces a novel hierarchical quality cue organization perspective for long-term multimodal AQA and proposes MLCR to progressively refine score-relevant cues across intra-modal representation, cross-modal interaction, and stage-wise aggregation levels.

	\item We design an intra-modal decoupling encoder that jointly captures global context and local frequency cues while preserving identity semantics, improving multi-granularity quality representation within each modality.
	
	\item We incorporate a fused state-conditioned cross-modal interaction strategy that dynamically mines uncovered incremental quality cues from modality discrepancy space and progressively aggregates them through stage-wise fusion, while suppressing redundant responses.

	\item Extensive experiments on the RG and Fis-V datasets demonstrate that MLCR achieves the best or second-best performance among action quality assessment methods in metric of Spearman's correlation and MSE, highlighting its robustness and effectiveness.

\end{itemize}

\section{Related Work}

\subsection{Action Quality Assessment} 

The core objective of AQA is to establish a mapping between action observations and expert scores. Early methods formulated AQA as an end-to-end score regression, extracting global action representations from videos with visual backbones and directly predicting scores via a regressor\cite{parmar2019and}. Treating expert scores as deterministic ground truth is a widely adopted simplifying assumption. This assumption is reasonably justified and facilitates model training, thereby establishing a foundational paradigm on short-term action benchmarks such as diving. Subsequent studies further recognized the scoring uncertainty arising from the subjective nature of expert evaluation\cite{tang2020uncertainty}, and introduced relative ranking learning, contrastive learning, or distribution-aware modeling\cite{doughty2019pros,yu2021group,li2022pairwise,bai2022action}, enabling models to learn not only the mapping between individual samples and scores but also the relative quality differences among performances.

Beyond holistic score regression, recent studies have explored finer-grained scoring formulations. Some approaches decompose action sequences into phases, steps, or technical elements, estimating local quality contributions before aggregating them into a final score\cite{xu2025human,xu2022finediving,xu2022likert}. Others incorporate scoring rubrics, action terminology, or textual feedback to embed expert knowledge into the assessment process, improving model interpretability\cite{dong2026uil,zhou2024cofinal}. These efforts suggest that AQA is not a simple regression task, but a quality modeling problem jointly governed by scoring criteria, action structure, and expert judgment logic. However, long-term actions such as rhythmic gymnastics and figure skating involve continuous choreography, element-level transitions, and rhythmic variation, while existing datasets often lack explicit procedural or element-level annotations.

This makes it difficult for existing methods to explicitly organize structurally coherent and score-relevant cues over long temporal sequences. In contrast, this paper proposes a quality representation framework aligned with expert scoring logic, progressively identifying and reinforcing score-relevant quality cues across three levels, namely intra-modal representation, cross-modal interaction, and stage-wise aggregation.

\subsection{Long-Term Temporal Modeling}  

Temporal modeling is fundamental to understanding long-term action quality. Early recurrent architectures, such as RNNs, LSTMs, and GRUs, accumulate temporal context through recursive hidden states, but suffer from vanishing gradients as sequences grow, limiting long-range dependency retention\cite{parmar2019action}. Transformer-based methods overcome this bottleneck via self-attention, enabling arbitrary temporal dependency modeling and strong global context capture\cite{gedamu2023fine}. However, their computational complexity grows quadratically with sequence length, limiting their efficiency in long-video scenarios. To address this issue, variants such as sparse attention and hierarchical transformers reduce computational cost through temporal downsampling or hierarchical aggregation\cite{zahan2024learning,wang2021tsa}. Nevertheless, while improving the efficiency of long-sequence modeling, they may weaken feature expressiveness by losing fine-grained temporal cues.

Recently, SSMs represented by S4~\cite{S4gu2022ICLR} and Mamba~\cite{gu2024mamba} have demonstrated advantages in long-sequence action understanding, owing to their linear computational complexity and long-range modeling capacity~\cite{wang2025learning}. By recursively compressing sequences through continuous-time state transitions, SSMs are well suited to capturing smooth global temporal trends. However, existing methods typically rely on a single operator to process all temporal cues, encoding both long-term execution patterns and short-term variations into a shared representation space. This unified paradigm tends to entangle quality cues across granularities, undermining simultaneous preservation of global coherence and local detail sensitivity. For SSMs, the recurrent mechanism attenuates high-frequency dynamics, hindering retention of sparsely distributed local quality details. Frequency analysis offers a complementary perspective for local temporal modeling\cite{quan2023mawkdn}. Recent work introduces frequency-domain cues into AQA to capture subtle differences between clips~\cite{wang2025adaptive}. However, it enhances temporal representations from a single frequency-domain perspective, without modeling the multi-granularity structural differences between global and local cues.

Existing long-term AQA methods typically rely on a single encoder for multi-scale quality modeling, conflating global trends with local perturbations. In contrast, we organize intra-modal cues via global temporal, local frequency, and identity branches to preserve long-range context, subtle quality variations, and modality semantics, yielding structured candidate cues for subsequent multimodal interaction.

\subsection{Multimodal Fusion}

Multimodal fusion integrates heterogeneous information from distinct semantic spaces. According to the fusion mechanism, existing methods can be divided into three categories. Direct fusion includes feature concatenation, weighted summation, and prediction-level ensembling. It combines modality representations with low complexity but struggles to distinguish shared information, modality-specific cues, and noise\cite{xia2023skating}. Beyond direct fusion, subsequent studies mainly improve fusion through cross-modal interaction and structured modality selection perspectives. Interaction-based fusion employs cross-modal attention or transformers to establish token-level, temporal, or semantic correspondence between modalities, enhancing modality consistency and information exchange\cite{xu2025quality,wang2025attention}. Structured or adaptive fusion includes shared-specific representation learning, gating, and mixture-of-experts (MoE) methods. These methods reduce redundancy and alleviate modality imbalance through representation decomposition, modality selection, or expert assignment\cite{zeng2024multimodal,xu2026mcmoe}.

Despite these advances, most fusion strategies still regard the fused representation as a one-shot aggregation result of available modality features, or determine interaction weights according to inter-modal similarity, semantic relevance, and overall modality reliability. These methods usually organize multimodal information through cross-modal alignment or consensus enhancement~\cite{wang2025learning,wang2025attention}, implicitly assuming that higher inter-modal similarity is more beneficial for fusion. This may overlook complementary cues embedded in modality discrepancies. For multimodal AQA, complementary cues are particularly relevant because RGB, optical flow, and audio characterize action quality from different aspects, including body posture, motion dynamics, and action-music synchronization. Therefore, discriminative information does not always lie in highly similar regions across modalities, but also come from the uncovered incremental evidence.

Different from them, rather than fusing modality features directly or reinforcing cross-modal consensus, we dynamically mine incremental quality cues from the cross-modal discrepancy space conditioned on the current fused state, while suppressing redundant responses. Through stage-wise aggregation of modality-specific and cross-modal complementary evidence, the fused representation is progressively refined into a more discriminative long-term quality representation.

\section{Method}

\begin{figure*}[t]
	\centering
	\includegraphics[width=18cm]{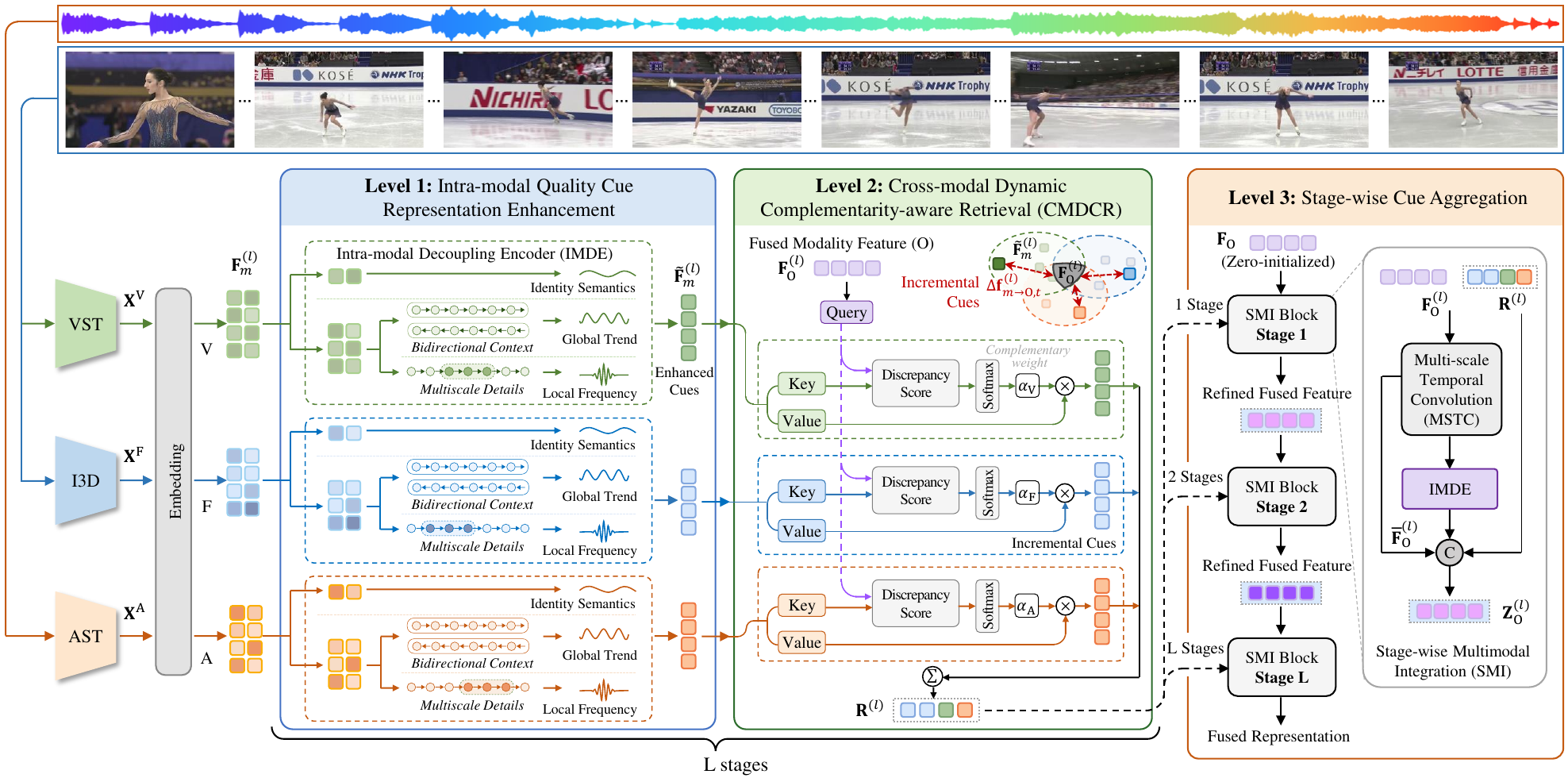}
	\caption{The architecture of MLCR. RGB, optical flow, and audio features are extracted by pretrained VST, I3D, and AST and projected into a shared latent space, with a zero-initialized fused feature serving as the initial cross-modal representation. MLCR refines quality cues at three levels. Level 1 uses IMDE to decouple each modality into identity semantics, global temporal context, and local frequency details. Level 2 employs CMDCR to retrieve discrepancy-guided complementary cues from modality-specific representations and enrich the fused state. Level 3 performs stage-wise cue aggregation through SMI blocks, where MSTC captures multi-scale temporal patterns and the fused representation is progressively updated.}
	\label{Fig_MLCR}
\end{figure*}

This section presents the proposed MLCR framework, as shown in Fig.\ref{Fig_MLCR}. Rather than treating long-term multimodal AQA as one-shot sequence regression, MLCR formulates it as a process of organizing and accumulating multi-level quality cues. It progressively builds discriminative multimodal representations for long-term action scoring through intra-modal cue decomposition, cross-modal complementary interaction, and stage-wise fusion.

\subsection{Problem Formulation and Overview}

For scoring-oriented AQA tasks such as gymnastics and figure skating, quality judgment depends on both continuous performance and sparse local cues, including action details, execution errors, and rhythm deviations. Thus, long-term multimodal AQA requires not only stronger feature extraction but also effective organization of score-relevant quality cues for score prediction.

Given an action sample partitioned into $T$ segments, let $\mathbf{X}^{m} \in \mathbb{R}^{T \times d_{m}}$ denote the segment-level feature sequence of modality $m$, where $m\in\mathcal{M}$ and $\mathcal{M}= \{\mathrm{V}, \mathrm{F}, \mathrm{A}\}$. Here, $\mathrm{V}$, $\mathrm{F}$, and $\mathrm{A}$ represent RGB, optical flow, and audio, respectively, and $d_{m}$ signifies the feature dimension of each modality. The objective of the model is to learn a non-linear mapping $\mathcal{F}_\Theta$ that transforms these multimodal sequences into a predicted quality score $\hat{y}$:
\begin{equation}
	\hat{y} = \mathcal{F}_\Theta (\mathbf{X}^\mathrm{V}, \mathbf{X}^\mathrm{F}, \mathbf{X}^\mathrm{A}),
\end{equation}
where $\mathcal{F}_\Theta(\cdot)$ represents the proposed MLCR model, and $\Theta$ represents its learnable parameters.

Owing to the discrepancies in statistical distributions and semantic spaces across the three modalities, a lightweight embedding mapping $\varepsilon_{m}(\cdot)$ is employed to project the original features $\mathbf{X}^{m}$ into a shared latent space. With $C$ denoting the channel dimension, the projection is formulated as:
\begin{equation}
	\mathbf{F}_m^{(0)} = \varepsilon_m(\mathbf{X}^m) \in \mathbb{R}^{C \times T}.
\end{equation}
In addition, we further construct a zero-initialized fused-modality feature $\mathbf{F}_{\mathrm{O}}=\mathbf{0}\in \mathbb{R}^ {C\times T}$ in the shared latent space to progressively aggregate information from RGB, optical flow, and audio modalities.

At the $l$-th stage, $\mathbf{F}_{\mathrm{O}}^{(l)}$ denotes the fused cue state that has been organized from previous stages. MLCR updates this state by integrating the current modality-specific representations ${\left\lbrace \mathbf{F}_m^{(l)}\right\rbrace}_{m\in\mathcal{M}}$, which can be abstractly formulated as:
\begin{equation}
	\mathbf{F}_{\mathrm{O}}^{(l+1)}
	=
	\mathcal{O}_{\Theta}^{(l)}
	\left(
	\mathbf{F}_{\mathrm{O}}^{(l)},
	{\left\lbrace \mathbf{F}_m^{(l)}\right\rbrace}_{m\in\mathcal{M}}
	\right),
\end{equation}
where ${\mathcal{O}}_{\Theta}^{(l)}(\cdot)$ denotes the quality cue organization function, instantiated by intra-modal cue decomposition, cross-modal complementary retrieval, and stage-wise multimodal integration, as detailed in the following subsections.

\subsection{Intra-modal Decoupling Encoder} 

\begin{figure}[t]
	\centering
	\includegraphics[width=8cm]{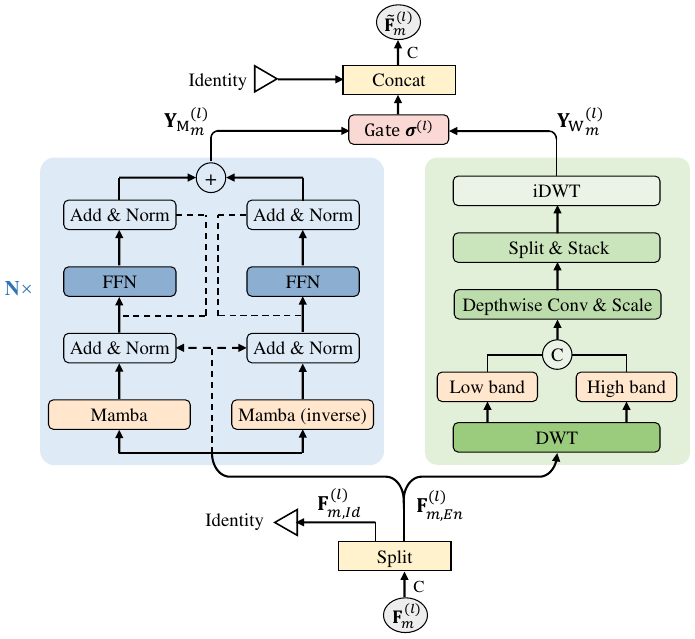}
	\caption{Illustration of IMDE. The input feature $\mathbf{F}_{m}^{(l)}$ is first split along the channel dimension into identity part and enhancement part. The identity part preserves modality-specific discriminative information, while the enhancement part uses global temporal and local frequency branches to capture long-range dependencies and local frequency variations, respectively. A learned gate $\sigma^{(l)}$ adaptively fuses the two enhanced features, which are then concatenated with the identity feature to obtain the output $\tilde{\mathbf{F}}_{m}^{(l)}$.} 
	\label{Fig_3_IMDE}
\end{figure}

In long-term AQA, each modality contains diverse quality cues that are often entangled. Using a single temporal operator may over-smooth local errors or insufficiently capture global execution patterns. IMDE therefore constructs a structured quality cue pool within each modality, as illustrated in Fig.\ref{Fig_3_IMDE}. IMDE splits features into quality-enhanced and identity-preserved parts. The former captures long-term dependencies via a global temporal branch and short-term perturbations via a local frequency branch, while the latter uses an identity branch to preserve modality-specific information with lower computational cost and mitigate semantic drift. IMDE provides structured candidate cues for subsequent cross-modal retrieval.

\smallskip
\textbf{Long-range Temporal Context Modeling}. 
Long-term action quality is typically assessed by jointly considering prior preparation, current execution, and subsequent completion effects. Unidirectional temporal modeling can only exploit contextual information from a single direction, limiting its ability to represent the bidirectional dependencies required for scoring. Therefore, we use a bidirectional state space branch for long-range temporal modeling. The forward Mamba captures historical dependencies, while the backward Mamba introduces future context, enabling the model to capture more complete global quality semantics with linear complexity.

Given the input feature $\mathbf{F}_m^{(l)} \in \mathbb{R}^{C \times T_l}$ of modality $m$ at the $l$-th ($l \in \{1,\ldots,L\}$) stage, we first split it along the channel dimension into two non-overlapping subsets:
\begin{equation}
	\text{Split}(\mathbf{F}_m^{(l)}, \lambda)=[\mathbf{F}_{m,\mathrm{Id}}^{(l)}; \mathbf{F}_{m,\mathrm{En}}^{(l)}]_C ,
\end{equation} 
where $\mathbf{F}_{m,\mathrm{En}}^{(l)} \in \mathbb{R}^{C_{\mathrm{En}}\times T_l}$ represents the part to be enhanced, and $\mathbf{F}_{m,\mathrm{Id}}^{(l)} \in \mathbb{R}^{C_{\mathrm{Id}}\times T_l}$ denotes the identity bypass. $[\cdot ;\cdot ]_C$ denotes concatenation along the channel dimension. $\lambda \in (0,1]$ is the split ratio, with $C_{\mathrm{Id}}=\left\lfloor \lambda C \right\rfloor$ and $C_{\mathrm{En}}=C- C_{\mathrm{Id}}$.

Let $\mathcal{M}_{f}^{(n)}$ and $\mathcal{M}_{b}^{(n)}$ denote the forward and backward SSM encoders in the $n$-th Bi-Mamba unit ($n=1,\dots,N$), respectively. The recursive update is formulated as:
\begin{equation}
	\resizebox{.95\linewidth}{!}{$
		\displaystyle
		\mathbf{F}_{m,\mathrm{En}}^{(l,n)}=
		\frac{1}{2}\left(
		\mathcal{M}_{f}^{(n)}\!\left(\mathbf{F}_{m,\mathrm{En}}^{(l,n-1)}\right)
		+\mathrm{Flip}^{-1}\!\left(
		\mathcal{M}_{b}^{(n)}\!\left(\mathrm{Flip}\!\left(\mathbf{F}_{m,\mathrm{En}}^{(l,n-1)}\right)\right)
		\right)
		\right),
		$}
\end{equation}
where $\mathrm{Flip}(\cdot)$ denotes temporal reversal. The final output of the temporal branch is $\mathbf{Y_M}_{m}^{(l)}=\mathbf{F}_{m,\mathrm{En}}^{(l,N)}$.

\smallskip
\textbf{Sparse Local Perturbation}.
Although recurrent state compression is effective for long-range temporal modeling, it may weaken sparse high-frequency quality cues. To compensate the global temporal branch, we introduce a lightweight Haar wavelet branch as a structural complement to retain short-term variations and high-frequency perturbations.

Given the enhanced feature $\mathbf{F}_{m,\mathrm{En}}^{(l)}$, the single-level Haar discrete wavelet transform (DWT) decomposes the temporal sequence into low-frequency trend components and high-frequency detail components.
The decomposed subbands are enhanced using depthwise temporal convolution and learnable channel scaling, yielding $\widetilde{\mathbf{L}}^{(l)}$ and $\widetilde{\mathbf{H}}^{(l)}$ as the refined low-frequency and high-frequency subbands, respectively, which are then reconstructed via inverse DWT to obtain the local frequency branch representation:
\begin{equation}
	{{\mathbf{Y}}_\mathrm{W}}_{m}^{(l)}=\left(\widehat{\mathbf{L}}^{(l)} \uparrow 2\right) * \mathbf{g}_{\mathrm{L}}
	+ \left(\widetilde{\mathbf{H}}^{(l)} \uparrow 2\right) * \mathbf{g}_{\mathrm{H}},
\end{equation}
where $\widehat{\mathbf{L}}^{(l)}$ denotes the reconstructed low-frequency representation, $\mathbf{g}_{\mathrm{L}}$ and $\mathbf{g}_{\mathrm{H}}$ denote the low-pass and high-pass reconstruction filters, respectively, and $\uparrow 2$ denotes upsampling along the temporal dimension.

\smallskip
\textbf{Adaptive Aggregation}. The global temporal branch and the local frequency branch provide different types of quality cues. The former captures overall action continuity and long-range rhythmic structures, whereas the latter captures short-term local perturbations and high-frequency details. Therefore, we design an adaptive gated fusion mechanism to dynamically balance their contributions according to the cue requirements at different temporal positions.

Let $\mathbf{f}_{m,\mathrm{En},t}^{(l)} \in \mathbb{R}^{C_{\mathrm{En}}}$ denote the feature vector at time step $t$. The scalar gate coefficient at time step t is computed as $\sigma_t^{(l)}
=\operatorname{Sigmoid} (\mathrm{MLP} (\mathbf{f}_{m,\mathrm{En},t}^{(l)}))$.
The adaptive fused representation at time step $t$ is defined as $\mathbf{y}_{m,\mathrm{En},t}^{(l)}=\sigma _t^{(l)} \mathbf{y}_{\mathrm{M},t}^{(l)}+\left(1-\sigma _{t}^{(l)}\right)\mathbf{y}_{\mathrm{W},t}^{(l)}$. The $\mathbf{y}_{\mathrm{M},t}^{(l)}$ and $\mathbf{y}_{\mathrm{W},t}^{(l)}$ denote the feature vectors at the $t$-th temporal position in ${\mathbf{Y}_\mathrm{M}}_{m}^{(l)}$ and ${\mathbf{Y}_\mathrm{W}}_{m}^{(l)}$, respectively. Accordingly, the adaptive gated fusion is:
\begin{equation}
	\mathbf{Y}_{\mathrm{gated}}^{(l)}=\boldsymbol{\sigma }^{(l)} \odot {\mathbf{Y}_\mathrm{M}}_{m}^{(l)} + \left(\textbf{1}-\boldsymbol{\sigma }^{(l)}\right)\odot {\mathbf{Y}_\mathrm{W}}_{m}^{(l)},
\end{equation}
where $\boldsymbol{\sigma }^{(l)} \in \mathbb{R}^{C_{\mathrm{En}}\times T_l}$ denotes the gating matrix obtained by broadcasting $\{\sigma_t^{(l)}\}_{t=1}^{T_l}$ along the channel dimension, $\odot$ denotes element-wise multiplication, and $\mathbf{1}$ is an all-one matrix of the same size.
The gated representation is then further processed with dropout and layer normalization, yielding the dual-branch fused output $\widetilde{\mathbf{Y}}_{\mathrm{MW}}^{(l)}$:
\begin{equation}
	\widetilde{\mathbf{Y}}_{\mathrm{MW}}\mathrm{}^{(l)}=\operatorname{LN}\!\left(\operatorname{Drop}\!\left(
	\mathbf{Y}_{\mathrm{gated}}^{(l)}\right)\right).
\end{equation}

Finally, the dual-branch time-frequency feature $\widetilde{\mathbf{Y}}_{\mathrm{MW}}^{(l)}$ is concatenated with the identity feature along the channel dimension, yielding the IMDE-enhanced output representation of modality $m$:
\begin{equation}
	\widetilde{\mathbf{F}}_{m}^{(l)}=\mathrm{IMDE}(\mathbf{F}_{m}^{(l)})=\mathrm{Concat} (\mathbf{F}_{m,\mathrm{Id}}^{(l)};\widetilde{\mathbf{Y}}_{\mathrm{MW}}^{(l)})
	\in \mathbb{R}^{C\times T_l},
\end{equation}
where $\mathrm{IMDE}(\cdot)$ denotes the proposed IMDE module, and $T_l$ denotes the temporal length at the $l$-th stage.

\subsection{Cross-modal Dynamic Complementarity-aware Retrieval}

In long-term multimodal AQA, RGB, optical flow, and audio capture action quality from distinct semantic spaces, providing both complementary and redundant evidence whose relative contributions shift as the fused representation $\mathbf{F}_{\mathrm{O}}^{(l)}$ accumulates information across stages. However, conventional cross-modal attention computes alignment scores based on inter-modal similarity, which systematically reinforces consensus regions already encoded in the fused state while suppressing modality-specific discriminative cues.

\smallskip
\textbf{Complementarity-aware retrieval criterion.}
We regard modality $m$ as providing complementary evidence to the current fused representation when it contains score-relevant information not yet captured by $\mathbf{F}_{\mathrm{O}}^{(l)}$, characterized by:
\begin{equation}
	I\!\left(\tilde{\mathbf{F}}_m^{(l)};\, y \;\middle|\; \mathbf{F}_{\mathrm{O}}^{(l)}\right) > 0,
	\label{eq:cmi}
\end{equation}
where $y$ denotes the quality score. A similarity-driven retrieval that attends to features well-aligned with $\mathbf{F}_{\mathrm{O}}^{(l)}$ tends to reinforce information already captured, leaving the conditional mutual information $I$ in~\eqref{eq:cmi} small. In contrast, attending to the \emph{discrepancy} between each modality and the current fused state expands the retrieval toward regions where incremental, score-relevant evidence is more likely to reside-precisely where $I(\cdot)$ is larger. The complementary value of each modality is therefore not a fixed property of the modality itself, but evolves with $\mathbf{F}_{\mathrm{O}}^{(l)}$. At early stage, when $\mathbf{F}_{\mathrm{O}}^{(l)}$ carries limited information, the module permits broad complementary contributions from all modalities. As $\mathbf{F}_{\mathrm{O}}^{(l)}$ evolves across stages, when $\mathbf{F}_{\mathrm{O}}^{(l)}$ has accumulated rich multimodal context, leave only fine-grained residual cues as truly incremental evidence.
Motivated by this, we design the cross-modal dynamic complementarity-aware retrieval (CMDCR) module, which formulates cross-modal retrieval as the recovery of a conditional incremental evidence $\Delta\mathbf{F}_{\mathrm{O}}^{(l)}$ that supplements $\mathbf{F}_{\mathrm{O}}^{(l)}$ with what it does not yet encode.

\smallskip
\textbf{Discrepancy-driven complementary weight.}
CMDCR realizes this conditional dynamic retrieval through a discrepancy-driven attention mechanism under score supervision. Specifically, the fused modality feature $\mathbf{F}_{\mathrm{O}}^{(l)} \in \mathbb{R}^{C \times T_l}$ at each time step $t$ serves as the query, while the IMDE-enhanced modality features $\left\{\widetilde{\mathbf{F}}_{m}^{(l)}\right\}_{m\in \mathcal{M}}$ serve as the keys and values. For the $h$-th attention head, features are projected into a low-dimensional subspace:
\begin{equation}
	\resizebox{.91\linewidth}{!}{$
		\displaystyle
		\mathbf{q}_{t}^{(l,h)}=\mathbf{W}_{\mathrm{Q}}^{(h)}\mathbf{f}_{\mathrm{O},t}^{(l)},\quad
		\mathbf{k}_{m,t}^{(l,h)}=\mathbf{W}_{\mathrm{K}}^{(h)}\widetilde{\mathbf{f}}_{m,t}^{(l)},\quad
		\mathbf{v}_{m,t}^{(l,h)}=\mathbf{W}_{\mathrm{V}}^{(h)}\widetilde{\mathbf{f}}_{m,t}^{(l)},
		$}
\end{equation}
where $\mathbf{W}_{\mathrm{Q}}^{(h)}$, $\mathbf{W}_{\mathrm{K}}^{(h)}$, and $\mathbf{W}_{\mathrm{V}}^{(h)}$ are learnable projection matrices. The discrepancy score is then computed as $e_{m,t}^{(l,h)}=-\frac{\mathbf{q}_{t}^{(l,h)} (\mathbf{k}_{m,t}^{(l,h)})^\top}{\sqrt{d_h}}$, where $d_h$ denotes the feature dimension of each attention head. Applying softmax across the modality dimension yields the normalized complementary weights:
\begin{equation}
	{\alpha_{m,t}^{(l,h)}}
	= \operatorname{Softmax}\!\left(\left\{e_{m,t}^{(l,h)}\right\}_{m\in\mathcal{M}}\right),
	\label{eq:softmax}
\end{equation}
which are assigned to RGB, optical flow, and audio modalities, respectively.
CMDCR assigns higher weight to modality features that differ from the current fused query, directing retrieval toward regions where the conditional mutual information in~\eqref{eq:cmi} is expected to be larger.

The conditional incremental cue at each time step $t$ is obtained by weighting modality-specific value features of modality $m$ with its complementary weight:
\begin{equation}
\Delta\mathbf{f}_{\mathcal{M}\rightarrow \mathrm{O},t}^{(l)}
=\sum_{m\in\mathcal{M}}	\Delta\mathbf{f}_{m\rightarrow \mathrm{O},t}^{(l)} 
= \sum_{m\in\mathcal{M}} \alpha_{m,t}^{(l)} \cdot \mathbf{v}_{m,t}^{(l)},
	\label{eq:delta_t}
\end{equation}
where $m\rightarrow \mathrm{O}$ denotes that the complementary cue is retrieved from modality $m$ to supplement the current fused representation $\mathrm{O}$. The complementary cue retrieved from all modalities is then obtained by aggregating these modality-specific contributions:
\begin{equation}
	\mathbf{R}^{(l)}
	=
	\operatorname{Stack}_{t=1}^{T_l}
	\left(
	\Delta\mathbf{f}_{\mathcal{M}\rightarrow \mathrm{O},t}^{(l)}
	\right)
	\in \mathbb{R}^{C\times T_l},
\end{equation}
where $\operatorname{Stack}(\cdot)$ arranges the retrieved cues along the temporal dimension.

It is worth noting that CMDCR does not treat all inter-modal discrepancies as complementary evidence. The discrepancy score expands the retrieval space beyond redundant consensus regions, while score supervision guides the value projection to emphasize quality-relevant cues. Task-irrelevant differences and unstructured noise are down-weighted during training, without estimating the mutual information term in~\eqref{eq:cmi}. Moreover, by operating on IMDE-refined features $\tilde{\mathbf{F}}_m^{(l)}$ rather than raw modality responses, CMDCR ensures that the discrepancy space reflects structured quality differences rather than noise.

\subsection{Stage-wise Discriminative Cue Fusion}   

Score-relevant cues in long-term multimodal AQA are often distributed across action phases and modalities. One-shot fusion compresses them into a single representation, making early fusion bias hard to correct and limiting complementary evidence integration. To address this issue, we model AQA as a stage-wise cue accumulation process, progressively supplementing missing information and building a more complete quality-aware representation across temporal scales.

\smallskip
\textbf{Multi-Scale Temporal Convolution}. 
Discriminative local quality variations often emerge at multiple temporal scales, such as short-term jitter, unstable continuous control, rhythm fluctuations, and the smoothness of transitions between consecutive movements. A single-scale convolution kernel is generally insufficient to capture such multi-granularity local dynamics. Therefore, we employ an MSTC module to enhance the modeling of local temporal patterns in the fused state. 
MSTC applies $K$ one-dimensional depthwise convolutions $\mathcal{F}_{k}^{\mathrm{dw}}(\cdot)$ with different kernel sizes along the temporal dimension, where the kernel set is defined as $\mathcal{K}=\{3,5,\dots,2K+1\}$. The resulting features are concatenated along the channel dimension and fused by a $1\times1$ Conv1D for channel compression and adaptive reweighting:
\begin{equation}
	\mathbf{U}  = \mathcal{F}_{1\times 1}\left(\text{Concat}_{k\in K} \mathcal{F}_{k}^{\mathrm{dw}}(\mathbf{F}_\mathrm{O}^{(l)} )\right).
\end{equation}
The final enhanced representation is obtained via residual aggregation and a lightweight feed-forward network:
\begin{equation}
	\bar{\mathbf{F}_\mathrm{O}^{(l)}} =\phi\!\left(\mathbf{F}_\mathrm{O}^{(l)}+\mathbf{U}+\mathrm{FFN}(\mathbf{F}_\mathrm{O}^{(l)}+\mathbf{U})\right),
\end{equation}
where $\phi(\cdot)$ denotes a nonlinear activation function.

\smallskip
\textbf{Stage-wise Multimodal Integration}.
After multi-scale temporal enhancement and cross-modal complementary retrieval, SMI updates the fused cue state at each stage. Specifically, at the $l$-th stage, MSTC first produces the locally enhanced fused representation $\bar{\mathbf{F}}_{\mathrm{O}}^{(l)}$, while CMDCR retrieves the complementary cue $\mathbf{R}^{(l)}$ from IMDE-enhanced modality features. Instead of directly concatenating raw modality features, SMI integrates the current fused state and the retrieved complementary cue to refine the stage-wise quality representation.

To further preserve the internal structure of the fused state, we also apply IMDE to $\bar{\mathbf{F}}_{\mathrm{O}}^{(l)}$, producing a time-frequency refined fused representation. The stage-wise aggregated representation is then formulated as:
\begin{equation}
	\mathbf{Z}_{\mathrm{O}}^{(l)} = \mathrm{Concat}(\bar{\mathbf{F}_\mathrm{O}^{(l)}};\mathbf{R}^{(l)};\mathrm{IMDE}(\bar{\mathbf{F}_\mathrm{O}^{(l)}})) \in \mathbb{R}^{3C \times T_l}.
\end{equation}
The concatenated representation is projected back to the original channel dimension through a lightweight transformation:
\begin{equation}
	\widehat{\mathbf{F}}_{\mathrm{O}}^{(l)}
	=
	\phi\!\left( \mathrm{BN}\!\left( \mathrm{Conv}_{1\times1}\!\left( \mathbf{Z}_{\mathrm{O}}^{(l)} \right) \right) \right),
\end{equation}
where $\phi$ denotes the GELU activation and $\mathrm{BN}$ denotes batch normalization. For the next stage, the updated fused representation and the IMDE-enhanced modality-specific representations are temporally downsampled as:
\begin{equation}
	\mathbf{F}_{\mathrm{O}}^{(l+1)}
	=
	\mathcal{P}_{l}
	\left(
	\widehat{\mathbf{F}}_{\mathrm{O}}^{(l)}
	\right),
	\quad
	\mathbf{F}_{m}^{(l+1)}
	=
	\mathcal{P}_{l}
	\left(
	\widetilde{\mathbf{F}}_{m}^{(l)}
	\right),
	\quad
	m\in\mathcal{M}.
\end{equation}
where $\mathcal{P}_l$ denotes the pooling operation. This formulation enables MLCR to progressively update the fused cue state while retaining modality-specific evidence for subsequent cross-modal retrieval. We stack three SMI blocks to balance cue accumulation and computational efficiency. After the final stage, the fused representation is fed into global average pooling and a regression head for score prediction, and the model is optimized with the mean squared error (MSE) loss.

\section{Experiments}
\subsection{Datasets and Evaluation Metric}
\textbf{Datasets.}
Experiments are conducted on two public long-term AQA benchmarks, including Rhythmic Gymnastics (RG) \cite{zeng2020hybrid} and Fis-V \cite{xu2020learning}. We follow the criteria established by the datasets and prior works.
The RG dataset includes 1000 videos covering four gymnastics routines, namely ball, clubs, hoop, and ribbon, with 250 videos for each routine. Each video is annotated with difficulty score, execution score, and total score, all given by judges according to official rules. The total score is the sum of the difficulty and execution scores, minus penalties. 
The Fis-V dataset consists of 500 videos from international ladies' singles short program figure skating competitions. Each video is approximately 2 minutes and 50 seconds long, with a frame rate of around 25 frames per second, capturing the entire performance of a single skater. All videos are annotated with the total element score and the total program component score.

\smallskip
\textbf{Evaluation Metric.}
In regression-based AQA studies \cite{zeng2020hybrid,xu2022likert,liu2022umt,su2020msaf,zeng2024multimodal}, \textbf{Spearman's rank correlation coefficient} (Sp. Corr.) $\rho$ and \textbf{MSE} are commonly used evaluation metrics. The former assesses the correlation between ground-truth and predicted scores, which ranges within $\rho  \in [-1,1]$, with values closer to 1 indicating stronger correlation. Given the rank sequences of the predicted and ground-truth scores, denoted as $\textbf{p} = (p_1,..., p_N)$ and $\textbf{q} = (q_1,..., q_N)$, respectively, Sp. Corr. is defined as follows:
\begin{equation}
	\rho =\frac{\sum_{i=1}^{N}\left ( p_{i} -\bar{p} \right ) \left ( q_{i}-\bar{q} \right ) }{\sqrt{\sum_{i=1}^{N} \left (p_{i} -\bar{p} \right )^{2} \sum_{i=1}^{N} \left ( q_{i}-\bar{q} \right )^{2} } } .
\end{equation}
Fisher's z-value \cite{parmar2019action} is employed to measure the average $\rho$ performance across actions.

\subsection{Implementation Details}

We implement MLCR in PyTorch and conduct both training and evaluation on an NVIDIA V100 GPU. The model takes temporal feature embeddings from three modalities, namely RGB, optical flow, and audio, as input. Specifically, RGB features are extracted by VST~\cite{liu2022video} pretrained on Kinetics-600~\cite{carreira2017quo}, optical flow features are extracted by I3D network~\cite{carreira2017quo} pretrained on Kinetics-400~\cite{carreira2017quo}, and audio features are extracted by audio spectrogram transformer (AST)~\cite{gong2021ast} pretrained on AudioSet~\cite{gemmeke2017audio}. The corresponding feature dimensions $d_{\mathrm{V}}$, $d_{\mathrm{F}}$, and $d_{\mathrm{A}}$ are 1024, 1024, and 768, respectively. To ensure temporal alignment across modalities, all modality sequences are unified to a fixed segment length during training via random cropping or zero padding, with the segment length set to 70 for the RG dataset and 130 for the Fis-V dataset. We use AdamW as the optimizer, with an initial learning rate of $8\times10^{-4}$, a weight decay of $1\times10^{-4}$, and a batch size of 64, while gradient clipping is applied. For the model configuration, the MSTC kernel parameter $K$ is set to 3, modality splitting is enabled with a ratio of 0.25, the number of 1D wavelet decomposition levels $q$ is set to 1. Different training epochs are adopted for different action categories. On the RG dataset, ball/clubs/hoop/ribbon are trained for 400/300/300/400 epochs, respectively. On the Fis-V dataset, both TES and PCS are trained for 300 epochs.

\begin{table*}
	\centering
	\footnotesize         
	\renewcommand\arraystretch{1.2}     
	\caption{Comparison with state-of-the-art approaches is measured by Spearman's rank correlation $\rho$ (the higher the better) and MSE (the lower the better) on the Rhythmic Gymnastics dataset. Fisher's z-value is used to compute the average $\rho$ across actions. Best and second-best results are highlighted in bold and underlined.}
	 
	\setlength{\tabcolsep}{2.2mm}
	{ 
		\scalebox{1} 
		{
			\begin{tabular}{lcclccccccccc}   
				
				\specialrule{1.5pt}{0pt}{0pt}

				\multirow{2}{*}{Method}  & \multirow{2}{*}{Par.(M)}  & \multirow{2}{*}{Features} & \multicolumn{5}{c}{Spearman Correlation($\uparrow$)} & \multicolumn{5}{c}{Mean Squared Error($\downarrow$)} \\
				\cmidrule(r){4-8}
				\cmidrule(r){9-13}
				& & & \multicolumn{1}{c}{Ball} & \multicolumn{1}{c}{Clubs} & \multicolumn{1}{c}{Hoop} & Ribbon &  \textbf{Avg.} & \multicolumn{1}{c}{Ball} & \multicolumn{1}{c}{Clubs} & Hoop & Ribbon & \textbf{Avg.}  \\
				\hline   
				
				C3D-SVR \cite{parmar2017learning}  & - & C3D & 0.357 & 0.551 & 0.495 & 0.516  & 0.483 & - & - & -&- & -  \\

				\hline
				
				MS-LSTM \cite{xu2020learning}  & 2.66 & I3D & 0.515 & 0.621 & 0.540 & 0.522 & 0.551 & 10.55 & 6.94 & 5.85 & 12.56 & 8.97\\
				
				ACTION-NET \cite{zeng2020hybrid}  & 28.08 & I3D + ResNet & 0.528 & 0.652 & 0.708 & 0.578  & 0.623 & 9.09 &6.40 & 5.93 & 10.23 & 7.91  \\
				
				GDLT \cite{xu2022likert}  & 3.16 & I3D & 0.553 & 0.720 & 0.712 & 0.562  & 0.644 & 8.78 & 6.25 & 6.02 & 9.39 & 7.61  \\
				
				HGCN \cite{zhou2023hierarchical}  & 0.50 & I3D  & 0.527 & 0.590 & 0.697 & 0.659 & 0.622 & 8.88 & 7.79 & 7.28 & 10.69 & 8.66\\
				
				CoFInAI \cite{zhou2024cofinal}  & 3.70 & I3D & 0.625 & 0.719 & 0.734 & 0.757 & 0.712 & 7.04 & 6.37 &5.81 &6.98 & 6.55 \\

				\hline
				MS-LSTM \cite{xu2020learning}  & 2.66 & VST & 0.621 & 0.661 & 0.670 & 0.695 & 0.663 & 7.52 & 6.04 & 6.16 & 5.78 & 6.37\\
				
				GDLT \cite{xu2022likert}  & 3.16 & VST & 0.746 & 0.802 & 0.765 & 0.741  & 0.765 & 5.90 & 4.34 & 5.70 & 6.16& 5.53  \\

				HGCN \cite{zhou2023hierarchical}  & 0.50 & VST  & 0.664 & 0.671 & 0.765 & 0.736 & 0.712 & 8.71 & 7.60 & 5.82 & 7.23 & 7.34 \\
				
				CoFInAI \cite{zhou2024cofinal}  & 3.70 & VST & \underline{0.809} & 0.806 & 0.804 & 0.810  & 0.807 & 5.07 & 5.19 & 6.37 & 6.30 & 5.73 \\
				
				VATP-Net \cite{gedamu2025visual} & 1.81 & VST & 0.800  &  0.810 &  0.780 & 0.769 & 0.790 & - & - &- & -& - \\

				\hline
				
				ACTION-NET \cite{zeng2020hybrid}  & 28.08 & VST + ResNet & 0.684 & 0.737 & 0.733 & 0.754  & 0.728 & 9.55 &6.36 & 5.56 & 8.15 & 7.41  \\

				UMT \cite{liu2022umt}  & 3.78 & VST + AST & 0.725 & 0.588 & 0.678 & 0.823  & 0.714 & - & - & - & - & -  \\
				
				
				PAMFN \cite{zeng2024multimodal}  & 18.06 & VST + AST + I3D  & 0.757 & 0.825 & \underline{0.836} & \underline{0.846} & 0.819 & 6.24  & 7.45 &5.21 &7.67 & 6.64  \\

				QGLV \cite{xu2025quality}  & 6.06  & VST + AST + CLIP  & \textbf{0.828} & \underline{0.827} & \underline{0.830} & 0.836 & 0.830 & \underline{4.83}  & \underline{4.30} & \textbf{4.77} & \underline{5.20} & \underline{4.78}  \\

				MLCR(Ours) & 25.89 & VST + AST + I3D  & 0.808 & \textbf{0.831} & \textbf{0.879} & \textbf{0.863} & \textbf{0.848} & \textbf{4.81}  & \textbf{4.01}  & \underline{4.80} & \textbf{4.50} & \textbf{4.53} \\			
				
				\specialrule{1pt}{0pt}{0pt}

			\end{tabular}%
		}    
		
		\label{tab_SOTA_RG} 
		\vspace{-5pt} 
	}
\end{table*}

\begin{table}
	\centering
	\footnotesize         
	\renewcommand\arraystretch{1.2}     
	\caption{Comparison with state-of-the-art approaches is measured by Spearman's rank correlation $\rho$ and MSE on Fis-V dataset.} 
	\setlength{\tabcolsep}{0.5mm}
	{ 
		\scalebox{0.945}
		{
			\begin{tabular}{lclcccccc}   
				\specialrule{1.5pt}{0pt}{0pt}

				\multirow{2}{*}{Method}  & \multirow{2}{*}{\shortstack[c]{Par.\\(M)}}  & \multirow{2}{*}{Features} & \multicolumn{3}{c}{Sp. Corr.($\uparrow$)} & \multicolumn{3}{c}{MSE($\downarrow$)} \\
				\cmidrule(r){4-6}
				\cmidrule(r){7-9}
				& & & \multicolumn{1}{c}{TES} & \multicolumn{1}{c}{PCS} & \textbf{Avg.} & \multicolumn{1}{c}{TES} & \multicolumn{1}{c}{PCS} & \textbf{Avg.}  \\
				\hline   
				
				C3D-LSTM \cite{parmar2017learning}  & - & C3D & 0.290 & 0.510 & 0.406 & 39.25 & 21.97 & 30.61  \\ 
				
				MS-LSTM \cite{xu2020learning} & 2.66 & C3D & 0.650 & 0.780 & 0.721 & 19.91 & 8.35 & 14.13 \\

				HGCN \cite{zhou2023hierarchical}  & 0.50 & VST & 0.246 & 0.221 & 0.234 & - & - & -\\
				
				TPT \cite{bai2022action}  & 11.82 & VST  & 0.570 & 0.760 & 0.676 & 27.50 & 11.25 & 19.38\\
				
				CoRe \cite{yu2021group}  & 2.51 & VST & 0.660 & 0.820 & 0.751 & 23.50 & 9.25 & 16.38\\
				
				GDLT \cite{xu2022likert}  & 3.16 & VST & 0.685 & 0.820 & 0.761 & 20.99 & 8.75 & 14.87  \\
				
				CoFInAI \cite{zhou2024cofinal}  & 5.24 & VST & 0.716 & 0.843 & 0.788 & 20.76 & 7.91 & 14.34 \\

				VATP-Net \cite{gedamu2025visual}  & 1.81 & VST  & 0.702 & 0.863 & 0.796 & - & - & - \\
				
				MLP-Mixer \cite{xia2023skating}  & 14.32 & VST+AST & 0.680 & 0.820 & 0.759 & 19.57 & 7.96 & 13.77\\
				
				UMT \cite{liu2022umt}  & 3.78 & VST+AST  & 0.716 & 0.822 & 0.774 & - & - & -  \\
				
				QGLV \cite{xu2025quality}  & 6.06 & VST+AST+CLIP    & 0.727  & 0.868 & 0.809 & \textbf{18.65}  & \textbf{6.50} & \textbf{12.58}  \\
				
				PAMFN \cite{zeng2024multimodal}  & 18.06 & VST+AST+I3D  & \textbf{0.754}  & \underline{0.872} & \textbf{0.822} & 22.50  & 8.16 & 15.33  \\

				MLCR(Ours)  & 25.89 &  VST+AST+I3D  & \underline{0.744}  & \textbf{0.877}  & \underline{0.821} & \underline{19.09}  & \underline{7.33}  & \underline{13.21} \\			
				
				\specialrule{1pt}{0pt}{0pt}

			\end{tabular}%
		}    
		\label{tab_SOTA_Fis-v} 
		\vspace{-5pt} 
	}
\end{table}

\subsection{Comparisons with State-of-the-arts}

\textbf{Comparisons on the Rhythmic Gymnastics dataset.}
The RG dataset features complex human-apparatus interactions and pronounced long-range temporal dependencies, posing challenges to long-term cue organization and cross-modal quality representation. Table~\ref{tab_SOTA_RG} compares MLCR with existing methods on RG.
MLCR achieves the best overall performance, with an average Sp. Corr. of 0.848 and an average MSE of 4.53, surpassing both PAMFN\cite{zeng2024multimodal} and QGLV\cite{xu2025quality}.
Under the same multimodal input setting as PAMFN, MLCR improves the average Sp. Corr. from 0.819 to 0.848 and reduces the average MSE from 6.64 to 4.53, indicating that the gains mainly come from multi-level cue refinement rather than simply using more modalities.
For individual apparatus categories, MLCR achieves the best Sp. Corr. on clubs, hoop, and ribbon, reaching 0.831, 0.879, and 0.863, respectively. Notably, although MLCR (0.808) yields a slightly lower Sp. Corr. than QGLV (0.828) on ball, it achieves a better MSE (4.81 of MLCR and 4.83 of QGLV). One possible reason is that quality differences in ball depend more on apparatus control and local visual details, where strong visual-semantic priors may benefit ranking correlation, whereas MLCR maintains lower prediction error through stage-wise cue accumulation. Overall, MLCR achieves higher ranking consistency and lower prediction error across most categories, showing its effectiveness in organizing long-term multimodal cues.
Compared with early single-visual models such as C3D-SVR and MS-LSTM, VST-based methods improve performance but still struggle with rhythm variation, apparatus dynamics, and cross-modal synchronization. In contrast, MLCR jointly organizes appearance, motion, and rhythm cues from RGB, optical flow, and audio, providing a more complete basis for quality assessment and achieving superior performance on RG.

\smallskip
\textbf{Comparisons on the Fis-V dataset.}
Table~\ref{tab_SOTA_Fis-v} reports the comparison of MLCR with existing methods on the Fis-V dataset. Overall, MLCR demonstrates competitive performance with both the average Sp. Corr. and MSE reaching best or second-best levels among current SOTA methods. Specifically, on PCS, MLCR achieves the highest Sp. Corr. of $0.877$, surpassing QGLV ($0.868$), PAMFN ($0.872$), and VATP-Net ($0.863$), which indicates that the proposed method is more effective at capturing program-component-related cues such as choreography, expressiveness, and execution quality. On TES, MLCR achieves the second-best result of $0.744$ and remains competitive with QGLV, which uses additional vision-language representations. Compared with PAMFN, the superior performance of MLCR more directly reflects the effectiveness of the proposed model architecture under a comparable input setting. In particular, MLCR improves the PCS Sp. Corr. from $0.872$ to $0.877$ and reduces the average MSE from $15.33$ to $13.21$. Although MLCR achieves an average Sp. Corr. close to that of PAMFN, it shows clearer advantages in program-component scoring and overall error control, further demonstrating the effectiveness of its quality cue organization mechanism.

\begin{table*}
	\centering
	\footnotesize         
	\renewcommand\arraystretch{1.2}    
	\caption{Ablation studies on different component combinations of proposed IMDE in the Rhythmic Gymnastics dataset.} 
	\setlength{\tabcolsep}{2.3mm}
	{ 
		\scalebox{1}
		{
			\begin{tabular}{lccccccccccccc}   
				\specialrule{1.5pt}{0pt}{0pt}
				
				\multirow{2}{*}{Setting} & \multirow{2}{*}{Split} & \multirow{2}{*}{Bi-Mamba} & \multirow{2}{*}{Wavelet} & \multicolumn{5}{c}{Spearman Correlation($\uparrow$)} & \multicolumn{5}{c}{Mean Squared Error($\downarrow$)} \\
				\cmidrule(r){5-9}
				\cmidrule(r){10-14}
				&  &  &  & Ball & Clubs & Hoop & Ribbon &  \textbf{Avg.} & Ball & Clubs & Hoop & Ribbon & \textbf{Avg.}  \\
				\hline   

				no split & \ding{55}  &  \checkmark & \checkmark  & 0.695 & 0.791 & 0.804 & 0.809 & 0.778 & 7.31 & 4.25  & 4.86 & 5.78 & 5.55 \\
				
				split (Identity-only) & \checkmark & \ding{55}  & \ding{55} & 0.697 & 0.806 & 0.798 & 0.806 & 0.780 & 8.21 & 5.55  & 6.61 & 5.44 & 6.45\\ 
				
				split + Bi-Mamba & \checkmark &  \checkmark & \ding{55} & 0.739 & 0.815 & 0.847 & 0.823 & 0.809 & 6.67 & 4.40  & 4.92 & 5.05 & 5.26\\	
				
				split + Wavelet & \checkmark & \ding{55}  & \checkmark & 0.720 & 0.822 & 0.791 & 0.805 & 0.787 & 7.38 & \textbf{3.94}  & 5.72 & 6.02 & 5.77\\ \hline
				
				
				MLCR (Full)  & \checkmark & \checkmark  & \checkmark & \textbf{0.808} & \textbf{0.831} & \textbf{0.879} & \textbf{0.863} & \textbf{0.848} & \textbf{4.81} & 4.01  & \textbf{4.80}  & \textbf{4.50} & \textbf{4.53} \\

				\specialrule{1pt}{0pt}{0pt} 
				
			\end{tabular}%
		}    
		\label{tab_Ablation_RG} 
		\vspace{-5pt} 
	}
\end{table*}

\begin{table}
	\centering
	\footnotesize         
	\caption{Ablation studies on different component combinations of proposed IMDE in the Fis-V dataset.}         
	\renewcommand\arraystretch{1.2}    
	\setlength{\tabcolsep}{1.9mm}
	{
		\scalebox{1}
		{
			\begin{tabular}{lcccccc}  
				\specialrule{1.5pt}{0pt}{0pt}
				
				\multirow{2}{*}{Setting}  & \multicolumn{3}{c}{Sp. Corr.($\uparrow$)} & \multicolumn{3}{c}{MSE($\downarrow$)} \\
				\cmidrule(r){2-4}
				\cmidrule(r){5-7}
				& TES & PCS &  \textbf{Avg.} & \multicolumn{1}{c}{TES} & \multicolumn{1}{c}{PCS} & \textbf{Avg.}  \\
				\hline   

				no split  & 0.722 & 0.872  & 0.810 & \textbf{17.61} & \textbf{6.41} & \textbf{12.01} \\	
				
				split (Identity-only)  & 0.726 & 0.872  & 0.811 & 22.55 & 8.55 & 15.55 \\ 
				
				split + Bi-Mamba  & 0.726 & 0.862  & 0.816 & 22.95 & 9.01 & 15.98 \\	
				
				split + Wavelet  & 0.726 & 0.866  & 0.807 & 21.74 & 10.41 & 16.08 \\ 
				
				\hline
				
				MLCR (Full)   & \textbf{0.744} & \textbf{0.877}  & \textbf{0.821} & 19.09 & 7.33 & 13.21 \\ 
				
				\specialrule{1pt}{0pt}{0pt}  
				
			\end{tabular}%
		}    
		\label{tab_Ablation_FisV} 
		\vspace{-5pt} 
	}
\end{table}

\subsection{Ablation Study}

To analyze the role of different quality cue organization components in MLCR, we conduct ablation studies by removing or replacing key modules. All ablation experiments are performed on the RG and Fis-V datasets. We focus on the effects of intra-modal cue decoupling on representation learning, discrepancy-guided retrieval on cross-modal interaction, and stage-wise cue accumulation on final scoring performance.

\smallskip
\textbf{Effectiveness of IMDE}. 
Long-term AQA requires both global temporal semantics and local perturbation details, while preserving stable modality-specific information. To verify intra-modal cue decomposition, we compare variants with or without feature splitting, the Bi-Mamba global temporal branch, and the wavelet local frequency branch. Tables~\ref{tab_Ablation_RG} and \ref{tab_Ablation_FisV} report the results on the RG and Fis-V datasets, respectively. 
On RG, the full MLCR achieves the best overall performance, with an average Sp. Corr. of 0.848 and the lowest average MSE of 4.53, showing that combining identity semantics, global temporal cues, and local frequency cues enhances long-term quality representation. Removing either the identity branch or the enhancement branches leads to performance degradation, while using the split alone brings only limited gains, indicating that feature splitting itself is insufficient to organize quality cues effectively. In contrast, adding the Bi-Mamba global temporal branch produces more evident gains, confirming the importance of long-range context for fluency and rhythmic structure in rhythmic gymnastics. Further incorporating the wavelet local frequency branch yields the best results, showing that global evidence and local perturbation evidence are complementary.
On Fis-V, the full MLCR also achieves the highest average Sp. Corr. (0.821), although the no-split variant obtains the lowest MSE. This suggests that IMDE improves ranking consistency and fine-grained quality discrimination, while the stronger representation capacity may slightly enlarge absolute errors for a few samples. Nevertheless, its MSE remains competitive.

\smallskip
\textbf{Effectiveness of IMDE Across Different Backbones}.
To isolate the effect of IMDE from multimodal fusion, we conduct component-wise ablations with I3D and VST under single-backbone settings. As shown in Table~\ref{tab_Ablation_RG_backbone} and Table~\ref{tab_Ablation_FisV_backbone}, the full setting consistently achieves the best Sp. Corr. on RG and Fis-V. Although not always the lowest in MSE, its performance remains close to the best, indicating that ranking gains do not incur substantially larger absolute errors.
These results suggest that IMDE does not rely on a specific backbone and consistently enhances visual representation and temporal modeling, showing strong generalizability and plug-and-play capability. Moreover, applying split alone does not lead to stable performance gains, whereas introducing global temporal branch or local frequency branch after split brings more consistent improvements. Taking the I3D backbone on RG as an example, the average Sp. Corr. of identity is only 0.699, rising to 0.714 with Bi-Mamba and 0.722 with the local frequency branch. A similar trend is observed on Fis-V. This indicates that feature splitting mainly provides a basis for disentanglement, while the subsequent structured modeling of global semantics and local perturbations is the key factor that improves representation ability.

\begin{table*}
	\centering
	\footnotesize        
	\renewcommand\arraystretch{1.2}     
	\caption{Evaluation of IMDE with different backbone networks and component configurations on the Rhythmic Gymnastics dataset, measured by Spearman's rank correlation ($\uparrow$) and mean squared error ($\downarrow$).} 
	\setlength{\tabcolsep}{2.04mm}
	{ 
		\scalebox{1}
		{
			\begin{tabular}{lccccccccccccc}   
				\specialrule{1.5pt}{0pt}{0pt}
				
				\multirow{2}{*}{Setting} & \multirow{2}{*}{Split} & \multirow{2}{*}{Bi-Mamba} & \multirow{2}{*}{Wavelet} & \multicolumn{5}{c}{Spearman Correlation($\uparrow$)} & \multicolumn{5}{c}{Mean Squared Error($\downarrow$)} \\
				\cmidrule(r){5-9}
				\cmidrule(r){10-14}
				&  &  &  & Ball & Clubs & Hoop & Ribbon &  \textbf{Avg.} & Ball & Clubs & Hoop & Ribbon & \textbf{Avg.}  \\
				\hline   

				I3D + no split & \ding{55}  &  \checkmark & \checkmark  & 0.685 & 0.671 & 0.750 & 0.754 & 0.718 & 6.26 & \textbf{6.78}  & 6.32 & 7.65 & 6.75 \\
				
				I3D + split (Identity-only) & \checkmark & \ding{55}  & \ding{55} & 0.612 & 0.687 & 0.730 & 0.754 & 0.699 & 9.83 & 7.92 & 7.84 & 7.39 & 8.25 \\
				
				I3D + split + Bi-Mamba & \checkmark &  \checkmark & \ding{55} & 0.663 & 0.684 & 0.740 & 0.757 & 0.714 & 6.52 & 6.85 & 6.68 & \textbf{6.92} & 6.74 \\
				
				I3D + split + Wavelet & \checkmark & \ding{55}  & \checkmark & 0.671 & 0.692 & 0.759 & 0.756 & 0.722 & 6.38 & 8.15 & 7.64 & 8.03 & 7.55 \\ 
				
				\textbf{I3D (Full)} & \checkmark & \checkmark  & \checkmark & \textbf{0.722} & \textbf{0.698} & \textbf{0.775} & \textbf{0.758}  & \textbf{0.739}  & \textbf{5.65} & 7.13 & \textbf{6.26} & 7.21 & \textbf{6.56}  \\
				
				\hline
				
				VST + no split & \ding{55}  & \checkmark & \checkmark  & 0.721 & 0.783 & 0.788 & 0.710 & 0.753 & 7.02 & 4.83 & \textbf{4.53} & \textbf{6.90} & \textbf{5.82} \\
				
				VST + split (Identity-only) & \checkmark & \ding{55}  & \ding{55} & 0.695 & 0.774 & 0.783 & 0.729 & 0.747 & 8.30 & 6.00 & 6.86 & 10.47 & 7.91 \\
				
				VST + split + Bi-Mamba & \checkmark &  \checkmark & \ding{55} & 0.684 & 0.769 & 0.797 & 0.682 & 0.738 & 7.61 & \textbf{4.61} & 4.98 & 8.25 & 6.36 \\
				
				VST + split + Wavelet & \checkmark & \ding{55}  & \checkmark & 0.709 & 0.768 & 0.795 & 0.685 & 0.742 & \textbf{6.16} & 6.80 & 6.69 & 7.93 & 6.90 \\ 
				
				\textbf{VST (Full)} & \checkmark & \checkmark  & \checkmark  & \textbf{0.746} & \textbf{0.813} & \textbf{0.822} & \textbf{0.780}  & \textbf{0.792} & 7.46 & 4.91 & 4.60 & 7.23 & 6.05 \\

				\hline
			 
				\specialrule{1pt}{0pt}{0pt} 
				
			\end{tabular}%
		}    
		\label{tab_Ablation_RG_backbone} 
		\vspace{-5pt} 
	}
\end{table*}

\begin{table}
	\centering
	\footnotesize         
	\caption{Evaluation of IMDE with different backbone networks and component configurations on the Fis-V dataset.}         
	\renewcommand\arraystretch{1.2}     
	\setlength{\tabcolsep}{1.3mm}
	{
		\scalebox{1}
		{
			\begin{tabular}{lcccccc}   
				\specialrule{1.5pt}{0pt}{0pt}
				
				\multirow{2}{*}{Setting}  & \multicolumn{3}{c}{Sp. Corr.($\uparrow$)} & \multicolumn{3}{c}{MSE($\downarrow$)} \\
				\cmidrule(r){2-4}
				\cmidrule(r){5-7}
				& TES & PCS &  \textbf{Avg.} & \multicolumn{1}{c}{TES} & \multicolumn{1}{c}{PCS} & \textbf{Avg.}  \\
				\hline   
				
				I3D + no split  & 0.634 & 0.781  & 0.715 & 20.05 & 7.94 & 14.00 \\	
				
				I3D + split (Identity-only)  & 0.647 & 0.779  & 0.719 & 20.44 & 11.83 & 16.14 \\ 
				
				I3D + split + Bi-Mamba  & 0.637 & 0.790  & 0.722 & 26.36 & 7.99 & 17.18  \\	
				
				I3D + split + Wavelet  & 0.649 & 0.788  & 0.723 & 28.27 & 15.10 & 21.69   \\ 
				
				\textbf{I3D (Full)}  & \textbf{0.662} & \textbf{0.804} & \textbf{0.741} & \textbf{19.84} & \textbf{7.76} & \textbf{13.80}   \\
				
				\hline
				
				VST + no split  & 0.669 & 0.834  & 0.762 & \textbf{20.93} & 8.98 & \textbf{14.96} \\	
				
				VST + split (Identity-only)  & 0.696 & 0.828  & 0.772 & 29.29 & 12.73 & 21.01 \\ 
				
				VST + split + Bi-Mamba  & 0.696 & 0.838  & 0.777 & 21.48 & \textbf{8.83} & 15.16  \\	
				
				VST + split + Wavelet  & 0.704 & 0.837  & 0.779 & 30.35 & 16.14 & 23.25   \\ 
				
				\textbf{VST (Full)}  & \textbf{0.707} & \textbf{0.850}  & \textbf{0.784} & 21.00 & 9.44 & 15.22   \\ 
				
				\specialrule{1pt}{0pt}{0pt}  
				
			\end{tabular}%
		}    
		\label{tab_Ablation_FisV_backbone} 
		\vspace{-5pt} 
	}
\end{table}

\smallskip
\textbf{Effectiveness of CMDCR}.
Table \ref{tab_Ablation_Cross-modal} evaluates the effectiveness of CMDCR against cross-attention and uniform modality aggregation. CMDCR outperforms cross-attention, improving the average Sp. Corr. from 0.779 to 0.848 and reducing the average MSE from 5.69 to 4.53. This suggests that similarity-based selection may not always correspond to score relevance in long-term multimodal AQA, especially when useful evidence is distributed across modality-specific differences. Uniform aggregation slightly outperforms cross-attention, suggesting that avoiding similarity-biased selection partly alleviates redundancy. However, it treats all modalities equally and cannot adaptively adjust the retrieved cues according to the current fused state. In contrast, CMDCR achieves the best results across action categories, demonstrating that fused state-conditioned retrieval more effectively mines incremental scoring evidence from the cross-modal discrepancy space while suppressing irrelevant differences and redundant responses under score supervision. This highlights the importance of cross-modal dynamic complementary retrieval for the quality cue organization.

\begin{table}
	\centering
	\footnotesize              
	\renewcommand\arraystretch{1.2}     
	\caption{Ablation study of different cross-modal interaction strategies on the Rhythmic Gymnastics dataset.} 
	\setlength{\tabcolsep}{0.65mm}
	{
		\scalebox{0.9}
		{
			\begin{tabular}{lcccccccccc}   
				\specialrule{1.5pt}{0pt}{0pt}

				\multirow{2}{*}{Strategy} & \multicolumn{5}{c}{Spearman Correlation ($\uparrow$)} & \multicolumn{5}{c}{Mean Squared Error($\downarrow$)} \\
				\cmidrule(r){2-6}
				\cmidrule(r){7-11}
				& Ball & Clubs & Hoop & Ribbon &  \textbf{Avg.} & Ball & Clubs & Hoop & Ribbon & \textbf{Avg.}  \\
				\hline   

				Cross-attention & 0.740  & 0.794 & 0.777& 0.801 & 0.779  & 6.17 & 4.75 & 6.00 &5.83 & 5.69 \\
				Uniform         & 0.721  & 0.806 & 0.820 & 0.822 & 0.794  & 6.61 & 4.18 & \textbf{4.32} & 5.44 & 5.14 \\				
			
				CMDCR & \textbf{0.808} & \textbf{0.831} & \textbf{0.879} & \textbf{0.863} & \textbf{0.848} & \textbf{4.81} & \textbf{4.01} & 4.80 & \textbf{4.50} & \textbf{4.53} \\	
				
				\specialrule{1pt}{0pt}{0pt}  	
				
			\end{tabular}%
		}    
		\label{tab_Ablation_Cross-modal} 
		\vspace{-5pt} 
	}
\end{table}

\smallskip
\textbf{Effectiveness of Stage-wise Progressive Fusion}. 
Table~\ref{tab_Ablation_RG_Stage} compares the performance of the one-stage, two-stage, and three-stage settings on the RG dataset. 
Under gated fusion, the three-stage setting achieves the best average Sp. Corr. and the lowest average MSE, indicating that multi-stage updates can continuously supplement quality evidence that is insufficiently exploited in earlier stages and progressively correct early fusion bias. It should be noted that increasing fusion stages introduces richer hierarchical evidence but also higher computational cost. Thus, we adopt three stages to balance cues accumulation and efficiency.
Specifically, the one-stage setting compresses multimodal evidence into a final representation, which may overwhelm local critical evidence with global trends and makes it difficult to correct modality bias introduced during early fusion. The two-stage setting introduces intermediate cue updating, but its accumulation depth remains limited. The three-stage setting progressively integrates intra-modal specific cues and cross-modal complementary cues, forming a more complete long-term quality representation. This further confirms that long-term multimodal AQA requires multi-level stage-wise fusion to organize scoring cues across modalities, rather than processing each modality independently and merging them through concatenation.

\smallskip
\textbf{Effects of Gating Strategies}. Although multi-stage modeling provides richer hierarchical evidence, performance still depends on effective integration of cues from different sources. Table~\ref{tab_Ablation_RG_Stage} also compares three fusion strategies in IMDE, including sum, concat, and gated fusion. Under the all-stage setting, the average Sp. Corr. of sum and concat is 0.820 and 0.810, respectively, whereas gated substantially improves it to 0.848 and reduces the average MSE to 4.53, outperforming both alternatives. This suggests that with more fusion stages, cues from different stages, modalities, and temporal scales become more heterogeneous and redundant. Simple summation assumes equal contributions, while concatenation preserves more information but lacks an active mechanism for suppressing redundancy and noise. In contrast, gated fusion adaptively selects evidence sources according to the current stage and temporal position, thereby organizing multi-stage quality cues more effectively.

\begin{table}
	\centering
	\footnotesize         
	\renewcommand\arraystretch{1.2}    
	\caption{Results of different stage-wise modeling and fusion strategies on the Rhythmic Gymnastics dataset.} 
	\setlength{\tabcolsep}{0.5mm}
	{ 
		\scalebox{0.9}
		{
			\begin{tabular}{lccccccccccc}  
				\specialrule{1.5pt}{0pt}{0pt}

				\multirow{2}{*}{Strategy} & \multirow{2}{*}{Fusion} & \multicolumn{5}{c}{Spearman Correlation ($\uparrow$)} & \multicolumn{5}{c}{Mean Squared Error($\downarrow$)} \\
				\cmidrule(r){3-7}
				\cmidrule(r){8-12}
				& & Ball & Clubs & Hoop & Ribbon &  \textbf{Avg.} & Ball & Clubs & Hoop & Ribbon & \textbf{Avg.}  \\
				\hline   
				
				\multirow{3}{*}{One-stage}
				
				& sum  & 0.750  & 0.810  & 0.855  &  0.805 &  0.808 & 5.91 & 4.55 & 3.97 & 5.10 &4.88 \\
				& concat   & 0.750  & 0.813  & 0.846  &  0.812 & 0.808  & 6.24 & 4.89 & \textbf{4.05} & 5.08 &5.07\\
				& gated & 0.742 & 0.796 & 0.852 & 0.819 & 0.806 & 6.52 & 4.52 & 4.15 & 5.00 &5.05 \\		\hline
				
				\multirow{3}{*}{Two-stage} 
				& sum  & 0.775  &  \textbf{0.839} & 0.821  & 0.857  &  0.825 & 5.57 & 4.23 & 5.05 & \textbf{4.49} & 4.84\\
				& concat   & 0.749  & 0.807  & 0.846  &  0.839 & 0.813  & 6.05 & 4.14 & 4.25 & 4.75 &4.80\\
				& gated & 0.743 & 0.802 & 0.831 & 0.830 & 0.804 & 6.57 & 4.14 & 4.74 & 5.30 &5.19 \\	\hline
				
				\multirow{3}{*}{\shortstack{Three-stage\\(MLCR)}}  
				& sum  & 0.773  &  0.813 & 0.848  &  0.837 &  0.820 & 6.12 & 4.08 & 4.08 & 5.43 &4.93\\
				& concat   & 0.763  &  0.821 & 0.814  &  0.835 & 0.810  & 5.65 & 4.26 & 5.51 & 5.06 &5.12 \\
				& gated & \textbf{0.808} & 0.831 & \textbf{0.879} & \textbf{0.863} & \textbf{0.848} & \textbf{4.81} & \textbf{4.01} & 4.80 & 4.50 & \textbf{4.53} \\	
				
				\specialrule{1pt}{0pt}{0pt}  
				
			\end{tabular}%
		}    
		\label{tab_Ablation_RG_Stage} 
		\vspace{-5pt} 
	}
\end{table}

\smallskip
\textbf{Effects of Modality Combinations}. 
Table~\ref{tab_ablation_RG_RGB-Flow-Audio} analyzes modality contributions to quality cue organization on RG. The full three-modality MLCR achieves the best performance, with the highest average Sp. Corr. and lowest average MSE among all single-modal and bimodal settings, indicating that long-term AQA benefits from complementary multimodal cues.
Among single-modal settings, RGB performs best with an average Sp. Corr. of 0.792, followed by optical flow, while audio is much weaker with 0.339. This suggests that RGB-based pose appearance and human-apparatus relations are primary cues, while optical flow provides dynamic complements, and audio alone is insufficient for accurate scoring.
However, when combined with visual modalities, audio provides additional temporal structural information. Among bimodal settings, RGB+Flow performs best, with an average Sp. Corr. of 0.803 and MSE of 4.81. The audio-involved bimodal results show that audio brings limited improvement when combined with only one visual modality. In contrast, when RGB, optical flow, and audio are modeled jointly, the model can integrate appearance, motion, and rhythm evidence to form a more complete multimodal quality representation.

\begin{table*}
	\centering
	\footnotesize              
	\renewcommand\arraystretch{1.2}     
	\caption{Ablation study of different modality combinations on the Rhythmic Gymnastics dataset.} 
	\setlength{\tabcolsep}{2.7mm}
	{ 
		\scalebox{1}
		{
			\begin{tabular}{lccccccccccccc}   
				\specialrule{1.5pt}{0pt}{0pt}

				\multirow{2}{*}{Method}  & \multicolumn{3}{c}{Modalities}& \multicolumn{5}{c}{Spearman Correlation ($\uparrow$)} & \multicolumn{5}{c}{Mean Squared Error($\downarrow$)} \\
				
				\cmidrule(r){2-4}
				\cmidrule(r){5-9}
				\cmidrule(r){10-14}
				& RGB & Flow & Audio & Ball & Clubs & Hoop & Ribbon &  \textbf{Avg.} & Ball & Clubs & Hoop & Ribbon  & \textbf{Avg.}  \\
				\hline   
				
				\multirow{3}{*}{Single-modal}
				
				&   & \checkmark &   & 0.722 & 0.698 & 0.775 & 0.758  & 0.739  & 5.65 & 7.13 & 6.26 & 7.21 & 6.56  \\ 
				
				& \checkmark &   &   & 0.746 & 0.813 & 0.822 & 0.780  & 0.792 & 7.46 & 4.91 & 4.60 & 7.23 & 6.05  \\ 
				
				&   &   & \checkmark & 0.345 & 0.194 & 0.487 & 0.312  & 0.339  & 12.00 & 14.15 & 11.46 & 15.11 &13.18  \\ 
				
				\hline
				
				\multirow{3}{*}{Multimodal}  
				
				&  & \checkmark & \checkmark & 0.670 & 0.681 & 0.818 & 0.780 & 0.744  & 6.01 & 6.85 & 5.78 & 6.48 & 6.28\\
				
				& \checkmark &   & \checkmark & 0.681 & 0.781 & 0.811 & 0.710 & 0.750  & 8.25 & 5.30 & \textbf{4.15} & 7.39 & 6.27\\
				
				& \checkmark & \checkmark &  & 0.792 & 0.800 & 0.794 & 0.826 & 0.803  & 5.39 & 4.45 & 4.77 & 4.91 & 4.81\\
				
				\hline
				
				MLCR (Ours)   & \checkmark & \checkmark & \checkmark & \textbf{0.808} & \textbf{0.831} & \textbf{0.879} & \textbf{0.863} & \textbf{0.848}  & \textbf{4.81} & \textbf{4.01} & 4.80 & \textbf{4.50} & \textbf{4.53} \\			
				
				\specialrule{1pt}{0pt}{0pt}

			\end{tabular}%
		}    
		\label{tab_ablation_RG_RGB-Flow-Audio} 
		\vspace{-5pt} 
	}
\end{table*}

\begin{table}[t]
	\centering
	\footnotesize           
	\renewcommand\arraystretch{1.2}    
	\caption{Ablation studies of the split ratio $\lambda$ on the Rhythmic Gymnastics dataset.} 
	\setlength{\tabcolsep}{0.90mm}
	{ 
		\scalebox{0.94}
		{
			\begin{tabular}{lcccccccccc}   
				\specialrule{1.5pt}{0pt}{0pt}

				\multirow{2}{*}{\# $\lambda$} & \multicolumn{5}{c}{Spearman Correlation ($\uparrow$)} & \multicolumn{5}{c}{Mean Squared Error($\downarrow$)} \\
				\cmidrule(r){2-6}
				\cmidrule(r){7-11}
				& \multicolumn{1}{c}{Ball} & \multicolumn{1}{c}{Clubs} & \multicolumn{1}{c}{Hoop} & Ribbon &  \textbf{Avg.} & \multicolumn{1}{c}{Ball} & \multicolumn{1}{c}{Clubs} & \multicolumn{1}{c}{Hoop} & Ribbon & \textbf{Avg.}  \\
				\hline   
				
				$\lambda$=0.05 & 0.739 & 0.830 & 0.803 & 0.836 & 0.805 & 6.56 & 4.05 & 4.98 & 4.80 & 5.10 \\ 
				
				$\lambda$=0.25  & \textbf{0.808} & \textbf{0.831} & \textbf{0.879} & \textbf{0.863} & \textbf{0.848} & \textbf{4.81} & 4.01  & 4.80 & \textbf{4.50} & \textbf{4.53} \\ 
				
				$\lambda$=0.5 & 0.729 & 0.824 & 0.851 & 0.830 & 0.813 & 6.75 & \textbf{3.88}  & 4.61 & 5.04 & 5.07 \\ 
				
				$\lambda$=0.75 & 0.727 & 0.812 & 0.832 & 0.832 & 0.804 & 6.91 & 4.07 & \textbf{3.97} & 5.57 & 5.13 \\ 
				
				$\lambda$=0.95 & 0.754 & 0.774 & 0.836 & 0.805 & 0.794 & 5.63 & 4.47 & 4.72 & 6.27 & 5.27 \\

				\specialrule{1pt}{0pt}{0pt}  
				
			\end{tabular}%
		}    
		\label{tab_Ablation_RG_Split} 
		\vspace{-5pt} 
	}
\end{table}

\subsection{Hyperparameter and Efficiency Analysis}

\textbf{Split Ratio}.
To analyze the impact of the intra-modal split ratio $\lambda$, we conduct a hyperparameter study on the RG dataset, as shown in Table~\ref{tab_Ablation_RG_Split}. Different values of $\lambda$ have a clear impact on both Sp. Corr. and MSE, and $\lambda =0.25$ achieves the best overall performance. Since $\lambda$ controls the channel proportion assigned to the identity semantic branch, an overly small $\lambda$ may preserve insufficient original modality semantics, making the model more vulnerable to excessive transformation. Conversely, an overly large $\lambda$ compresses the channel capacity of the quality enhancement branch, weakening global temporal modeling and local frequency enhancement. Therefore, an appropriate split ratio provides a better balance between original information preservation and quality cue enhancement, producing more effective candidate cues for subsequent cross-modal retrieval and stage-wise fusion.

\begin{table}[t]
	\centering
	\footnotesize                
	\renewcommand\arraystretch{1.2}     
	\caption{Ablation study of different multi-kernel scales $K$ on the Rhythmic Gymnastics dataset.} 
	\setlength{\tabcolsep}{0.92mm}
	{ 
		\scalebox{0.96}
		{
			\begin{tabular}{lcccccccccc}   
				\specialrule{1.5pt}{0pt}{0pt}

				\multirow{2}{*}{\# K} & \multicolumn{5}{c}{Spearman Correlation ($\uparrow$)} & \multicolumn{5}{c}{Mean Squared Error($\downarrow$)} \\
				\cmidrule(r){2-6}
				\cmidrule(r){7-11}
				& \multicolumn{1}{c}{Ball} & \multicolumn{1}{c}{Clubs} & \multicolumn{1}{c}{Hoop} & Ribbon &  \textbf{Avg.} & \multicolumn{1}{c}{Ball} & \multicolumn{1}{c}{Clubs} & \multicolumn{1}{c}{Hoop} & Ribbon & \textbf{Avg.}  \\
				\hline   
				
				$K$=1 & 0.791 & 0.800 & 0.820 & 0.827 & 0.810 & 5.87 & 4.05 & 5.23 & 4.84 & 5.00 \\ 
				
				$K$=2 & 0.727 & 0.828 & 0.815 & 0.843 & 0.807 & 7.12 & 4.48 & \textbf{4.58} & 4.97 & 5.29 \\ 
				
				$K$=3  & \textbf{0.808} & \textbf{0.831} & \textbf{0.879} & \textbf{0.863} & \textbf{0.848} & \textbf{4.81} & \textbf{4.01}  & 4.80 & \textbf{4.50} & \textbf{4.53} \\ 
				
				$K$=4 & 0.723 & 0.826 & 0.828 & 0.831 & 0.806 & 6.37 & 4.45 & 4.72 & 5.11 & 5.16 \\ 	
				
				$K$=5 & 0.737 & 0.789 & 0.836 & 0.835 & 0.803 & 6.78 & 4.79 & 5.02 & 4.97 & 5.39 \\ 	
				
				\specialrule{1pt}{0pt}{0pt}  
				
			\end{tabular}%
		}    
		\label{tab_Ablation_RG_kernels} 
		\vspace{-5pt}
	}
\end{table}

\smallskip
\textbf{Multi-kernel Scales}.
To analyze the impact of the number of multi-scale temporal convolution kernels, we conduct experiments on the RG dataset with ($K$=1,2,3,4,5), as reported in Table~\ref{tab_Ablation_RG_kernels}. The best overall performance is achieved at ($K=3$), where both the average Sp. Corr. and the average MSE reach optimal values. These results suggest that the benefit of multi-scale modeling does not stem from simply increasing the number of scales, but rather from achieving a proper balance between representational capacity and redundancy control. Specifically, a single-scale setting ($K=1$) is insufficient to capture the multi-granularity discriminative cues in AQA, whereas increasing $K$ to 4 or 5 leads to consistent performance degradation, indicating that excessive scales may introduce redundant information and noise, weaken effective quality cue representation, and make optimization more difficult.

\begin{table}[t]
	\centering
	\footnotesize         
	\renewcommand\arraystretch{1.2}     
	\caption{Comparison of model complexity, inference latency, and performance of representative multimodal methods on the Rhythmic Gymnastics dataset.} 
	\setlength{\tabcolsep}{1mm}
	{ 
		\scalebox{1}
		{
			\begin{tabular}{llccccc}   
				\specialrule{1.5pt}{0pt}{0pt}

				Method & Features    & Params  &  Inference Time & Sp. Corr.   \\
				
				\hline   
				
				UMT \cite{liu2022umt}  &  VST + AST  & 3.78M     & 5ms  &  0.714    \\
				
				Joint-VA \cite{Badamdorj2021Joint}  &  VST + AST   & 1.97M    & 4ms  &  0.746    \\
				
				MSAF \cite{su2020msaf} &  VST + AST + I3D  & 5.56M     & 10ms  &  0.781    \\
				
				PAMFN \cite{zeng2024multimodal} & VST + AST + I3D   & 18.06M    & 33ms  &  0.819   \\
				
				MLCR (Ours) & VST + AST + I3D   & 25.89M   & 52ms  &  \textbf{0.848}   \\

				\specialrule{1pt}{0pt}{0pt}  
				
			\end{tabular}%
		}    
		\label{tab_Flops} 
		\vspace{-5pt} 
	}
\end{table}

\begin{figure}[t]
	\centering
	\includegraphics[width=8.2cm]{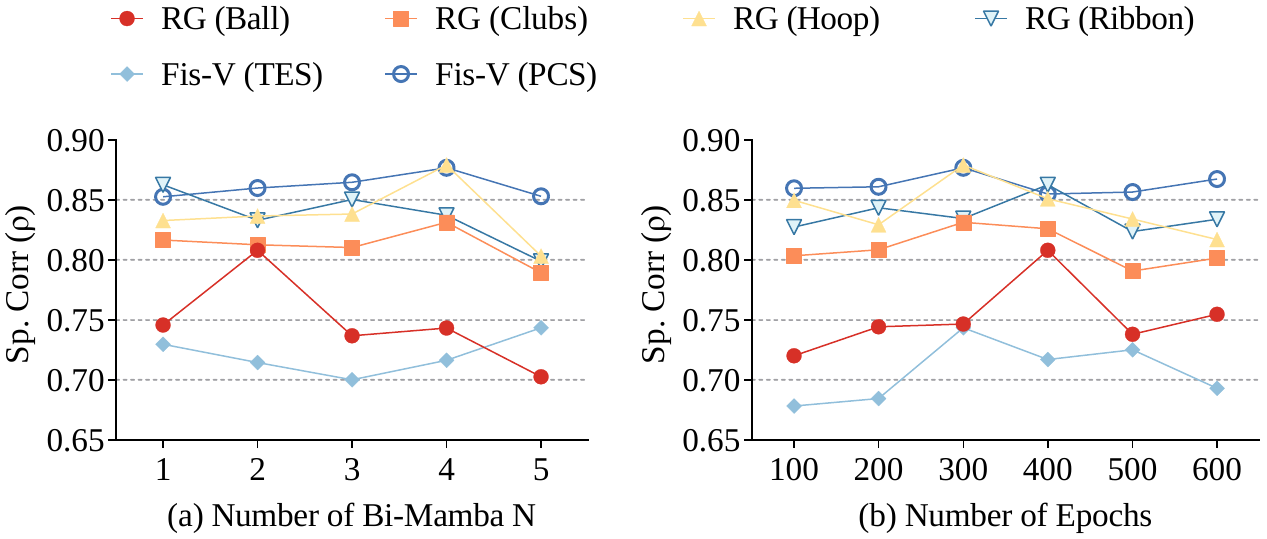}
	\caption{Hyperparameter analysis of the proposed MLCR on the Rhythmic Gymnastics and Fis-V datasets. (a) Evaluation of the number of Bi-Mamba units. (b) Evaluation of the number of training epochs.}
	\label{Fig_Number_of_fusionNet}
	\vspace{-5pt} 
\end{figure}

\begin{figure}[t]
	\centering
	\includegraphics[width=7cm]{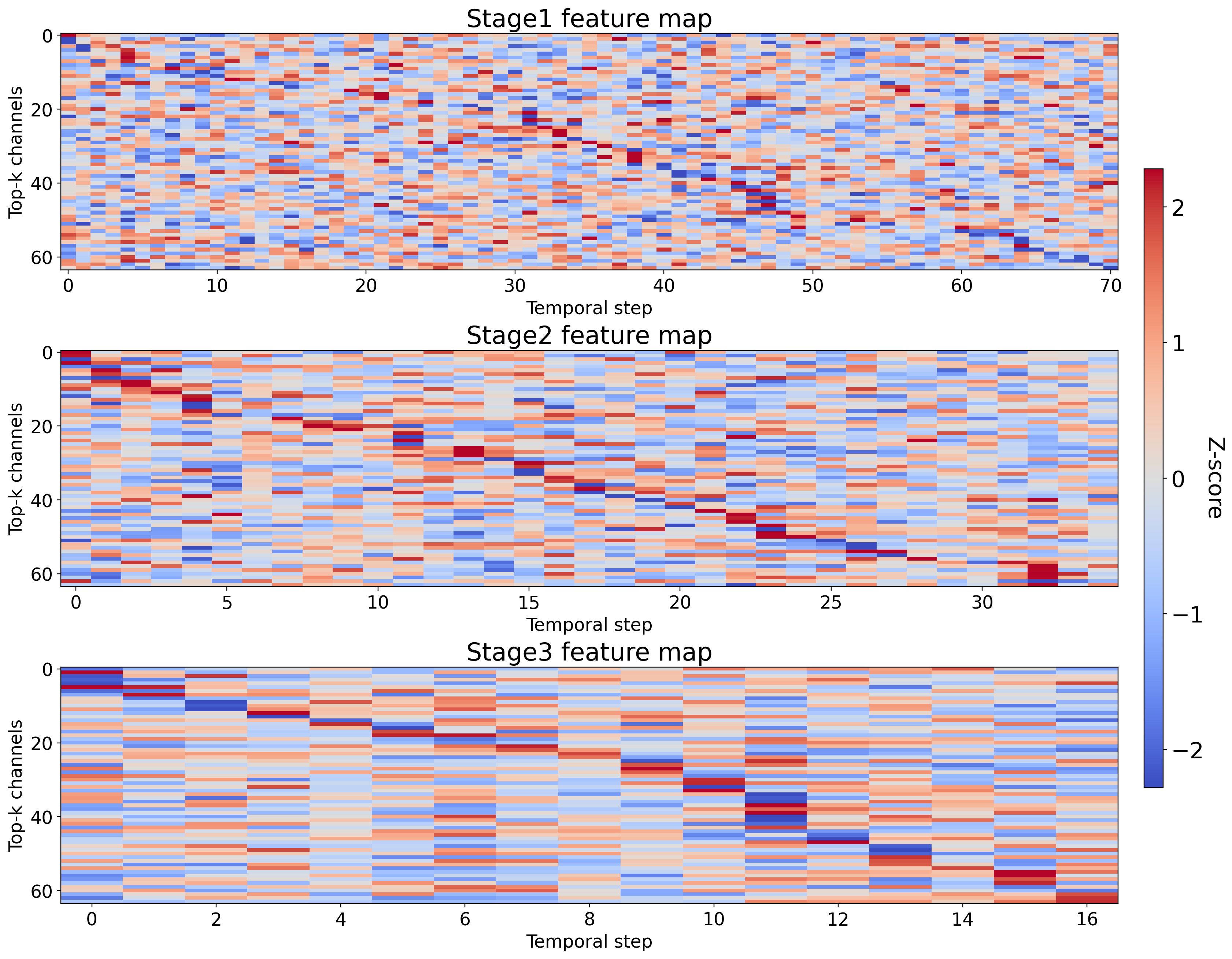}
	\caption{Visualization of intermediate feature response maps across three progressive fusion stages for a representative RG ball \#025 test sample.}
	\label{6_topk_heatmap}
	\vspace{-5pt}
\end{figure}

\smallskip
\textbf{Bi-Mamba Units}.
To analyze the impact of model depth on quality cue organization, we compare the performance obtained with different numbers of Bi-Mamba units ($N$=1,2,3,4,5). As shown in Fig.~\ref{Fig_Number_of_fusionNet}(a), the results on RG and Fis-V show clear task dependence and depth saturation, rather than a simple deeper-is-better trend. Specifically, ribbon reaches its best correlation of 0.863 at $N$=1, whereas clubs and hoop attain their optimal results at $N$=4 with 0.831 and 0.879, respectively. When $N$=5, performance drops markedly across all tasks. The reason is that although the action sequences are relatively long, an excessively deep multi-scan mechanism may dilute low-level fine-grained features during propagation and introduce issues such as state redundancy, noise propagation, and over-smoothed representations, weakening the model’s ability to capture subtle scoring cues. In addition, the increased parameter count further exacerbates the optimization difficulty under limited data.

\smallskip
\textbf{Training Strategies}.
Fig.~\ref{Fig_Number_of_fusionNet}(b) shows the Sp. Corr. variation under different training epochs on the RG and Fis-V datasets. The performance exhibits a clear stage-wise convergence pattern. In the early training phase, the correlation progressively improves as the number of epochs increases, indicating that the model gradually learns a more stable quality cue organization process. After a certain point, performance gradually saturates. 
Notably, the optimal number of epochs is not identical across different action categories, suggesting that different tasks exhibit varying sensitivity to training intensity. Based on this analysis, we set the training epochs to 400, 300, 300, and 400 for ball, clubs, hoop, and ribbon, respectively, and 300 epochs for both TES and PCS.

\smallskip
\textbf{Computational Efficiency Analysis}.
Table~\ref{tab_Flops} compares the efficiency and performance of different multimodal methods on RG. Although UMT\cite{liu2022umt} and Joint-VA\cite{Badamdorj2021Joint} based on RGB and audio modalities incur lower computational cost, their ability to capture complex spatiotemporal quality cues remains limited for fine-grained long-term tasks such as rhythmic gymnastics. Introducing optical flow captures richer motion dynamics and improves performance. 
Under the same trimodal input setting, MLCR achieves the highest Sp. Corr., showing that its gain mainly comes from multi-level quality cue organization rather than using more modalities. Although MLCR increases parameters and inference time to 25.89M and 52ms, the overhead remains acceptable for long-term AQA scenarios. The additional cost mainly comes from the bidirectional sequence encoding for long-term temporal refinement across stages, which introduces moderate overhead while improving the accuracy and stability of long-term quality representation. Overall, MLCR achieves a favorable trade-off between accuracy and computational efficiency.

\subsection{Visualization}

\textbf{Visualization of Stage-wise Feature Refinement}. 
Fig.~\ref{6_topk_heatmap} shows the intermediate feature responses of sample ball \#025 from the RG dataset across three progressive fusion stages. The horizontal axis denotes the temporal dimension, and the vertical axis denotes the top-$k$ channels with the largest temporal variance from the fused features at each stage, where $k=64$. The color intensity represents the normalized feature response magnitude.
The responses in stage 1 are scattered over time, indicating that the shallow fusion stage preserves rich local temporal details and is sensitive to short-term motion variations and fine-grained modality cues. In stage 2, high-response regions form clearer stripe-like and block-like patterns along the temporal axis, suggesting that the model starts to integrate cross-modal information more effectively over time. In stage 3, responses become compact and structured, indicating that the progressive fusion suppresses redundant activations and concentrates on discriminative temporal evidence. This visualization intuitively demonstrates the stage-wise evidence accumulation process of MLCR, moving from local detail perception to cross-modal integration and finally to discriminative evidence condensation.

\begin{figure*}[t]
	\centering
	\includegraphics[width=16cm]{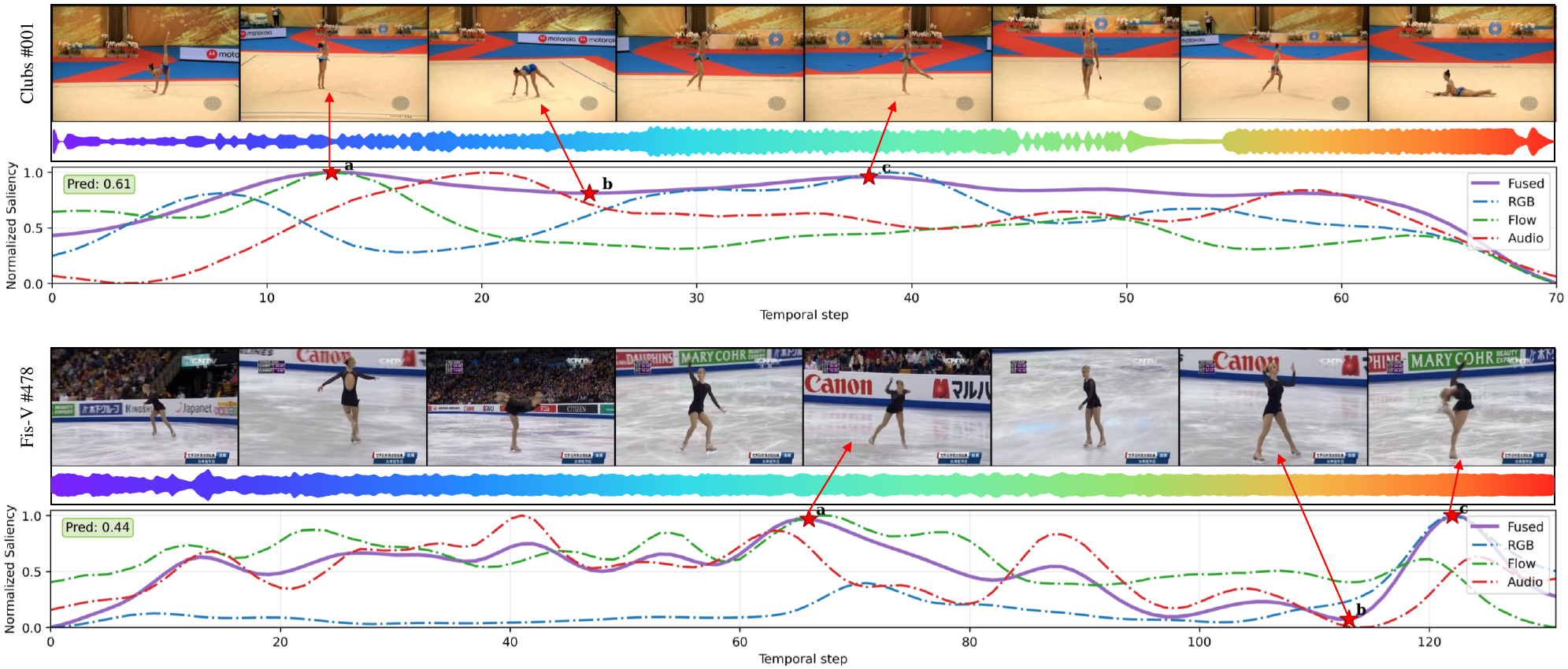}
	\caption{Qualitative visualization of multimodal temporal responses for test samples. The first sample is clubs \#001 from RG, and the second is video \#478 from the PCS class of Fis-V. The top row presents representative frames and corresponding audio, and the bottom row depicts the normalized saliency curves for RGB, optical flow, audio, and fused modalities. Higher curve values indicate greater contribution of corresponding time step to the final score.}
	\label{5_vis_Club_TES}
	\vspace{-5pt} 
\end{figure*}

\smallskip
\textbf{Visualization of Cross-modal Complementarity}. 
Fig.~\ref{5_vis_Club_TES} presents qualitative visualizations of multimodal temporal responses from samples in RG and Fis-V. The normalized temporal saliency curves reflect the contribution of different modalities at different temporal positions to the final prediction. The fused curve shows pronounced peaks only at a few critical intervals and remains low during transitions, indicating that MLCR locates discriminative quality cues rather than uniformly modeling the entire sequence.
Different modalities exhibit clear complementarity. The RGB modality shows trends broadly consistent with the fused curve peaks, suggesting strong sensitivity to body posture, apparatus relationships, and appearance cues. Optical flow is more sensitive to motion dynamics and action changes, providing evidence related to velocity, direction, and stability. Audio provides additional temporal cues related to rhythm perception, music structure, and action boundaries. These observations demonstrate that MLCR organizes complementary cues according to the assessment requirements of different temporal segments, rather than relying on a single modality or cross-modal consensus.

\begin{figure*}[t]
	\centering
	\includegraphics[width=18cm]{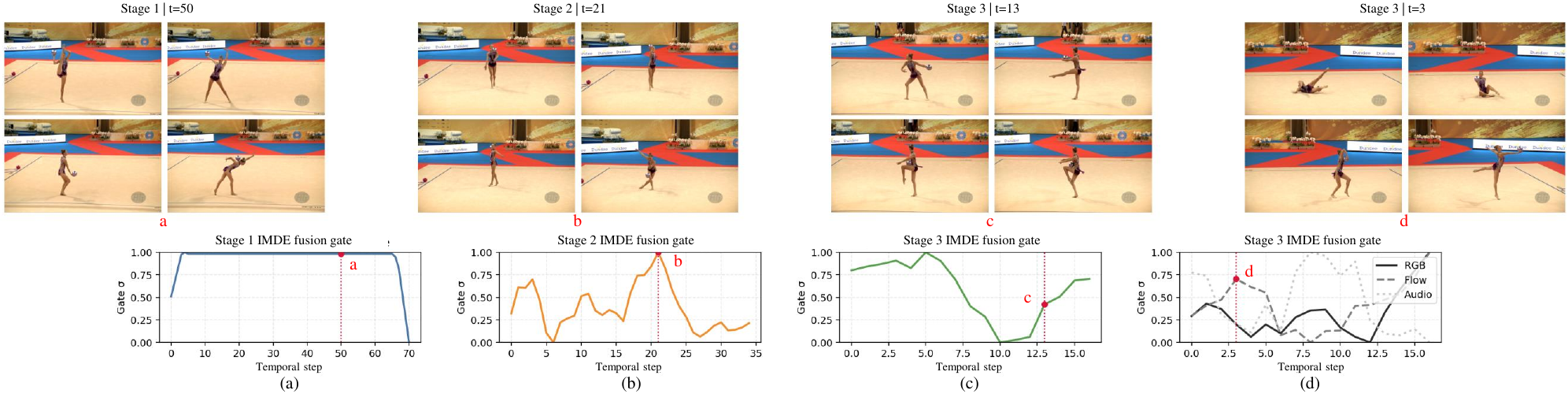}
	\caption{Qualitative visualization of IMDE gate responses for RG test sample ball \#033. The top row shows representative frames aligned with positions a–d. In the bottom row, (a)-(c) show fused modality gate responses at stages 1, 2, and 3, while (d) shows RGB, optical flow, and audio gate responses at stage 3.}
	\label{7_gate_RG}
	\vspace{-5pt} 
\end{figure*}

\smallskip
\textbf{Visualization of Adaptive IMDE Gate Responses}. 
Fig.\ref{7_gate_RG} visualizes the IMDE gating responses $\sigma$ on RG test sample ball \#033, including fused modality branch gates at stages 1-3 and RGB, flow, and audio branch gates at stage 3. Larger values indicate stronger preference for the global temporal branch, whereas smaller values indicate greater reliance on the local frequency branch.
IMDE dynamically adjusts the contributions of global and local evidence according to temporal position, fusion stage, and modality type. Specifically, stage 1 relies more on the global temporal branch over a long middle interval, suggesting that the shallow stage focuses more on integrating long-range context. Stage 2 increases reliance on the local frequency branch to better capture fine-grained variations. Stage 3 shifts from global structural modeling to local detail refinement, reflecting a transition from holistic understanding to fine-grained correction.
Different modalities also exhibit distinct preferences. Audio relies more on global temporal evidence in the middle segment for rhythmic structure, RGB prefers global temporal modeling at the beginning and end while focusing more on local frequency details in part of the middle segment, and optical flow increasingly relies on global temporal evidence in later segments. These results indicate that IMDE adaptively selects quality evidence across modalities and stages, improving the flexibility and discriminability of long-term multimodal AQA.

\section{Conclusion}

This paper addresses long-term multimodal action quality assessment by reformulating it as a process of quality cue organization and proposing MLCR, a multi-level cue refinement framework. Through intra-modal representation, cross-modal complementary interaction, and stage-wise cue fusion, MLCR progressively mines, filters, and integrates discriminative cues relevant to score prediction, thereby alleviating the degradation of key quality information caused by global temporal trends, modal redundancy, and one-shot fusion. Experiments on the Rhythmic Gymnastics and Fis-V datasets demonstrate that MLCR achieves competitive performance in both score correlation and prediction error, validating the effectiveness of the proposed quality evidence refinement mechanism. Future work will further explore more lightweight stage-wise fusion structures and extend this mechanism to more complex action assessment scenarios.

\bibliographystyle{IEEEtran}
\bibliography{IEEEabrv}

\vfill

\end{document}